**Huzaifa Mustafa Unjhawala**
Department of Mechanical Engineering,
University of Wisconsin-Madison,
Madison, WI 53706
e-mail: unjhawala@wisc.edu

**Ruochun Zhang**
Department of Mechanical Engineering,
University of Wisconsin-Madison,
Madison, WI 53706
e-mail: rzhang294@wisc.edu

**Wei Hu**
Department of Mechanical Engineering,
University of Wisconsin-Madison,
Madison, WI 53706
e-mail: whu59@wisc.edu

**Jinlong Wu**
Global Environmental Center,
California Institute of Technology,
Pasadena, CA 91125
e-mail: jinlong@caltech.edu

**Radu Serban**
Department of Mechanical Engineering,
University of Wisconsin-Madison,
Madison, WI 53706
e-mail: serban@wisc.edu

**Dan Negrut[1]**
Department of Mechanical Engineering,
University of Wisconsin-Madison,
Madison, WI 53706
e-mail: negrut@wisc.edu

# Using a Bayesian-Inference Approach to Calibrating Models for Simulation in Robotics

*In robotics, simulation has the potential to reduce design time and costs, and lead to a more robust engineered solution and a safer development process. However, the use of simulators is predicated on the availability of good models. This contribution is concerned with improving the quality of these models via calibration, which is cast herein in a Bayesian framework. First, we discuss the Bayesian machinery involved in model calibration. Then, we demonstrate it in one example: calibration of a vehicle dynamics model that has low degree-of-freedom (DOF) count and can be used for state estimation, model predictive control, or path planning. A high fidelity simulator is used to emulate the "experiments" and generate the data for the calibration. The merit of this work is not tied to a new Bayesian methodology for calibration, but to the demonstration of how the Bayesian machinery can establish connections among models in computational dynamics, even when the data in use is noisy. The software used to generate the results reported herein is available in a public repository for unfettered use and distribution.* [DOI: 10.1115/1.4062199]

## 1 Introduction

This contribution is motivated by the increasing use of simulation in the design of autonomous robots and vehicles. Simulation reduces design time, cuts costs, and leads to a safer development cycle and a more robust engineered solution [1]. For instance, one might be interested in choosing the model parameters that would allow simple two-degree-of-freedom (DOF) or eight-DOF vehicle models to be used effectively for model predictive control in an autonomy stack [2]. This poses a parameter calibration problem—data obtained from a vehicle is available and the task is to calibrate an expeditious model that is a good proxy for the dynamics of the actual vehicle. Traditionally, the parameter calibration problem for multibody systems has been posed as an optimization problem by coupling the integrator with an optimization procedure [3–7]. With the exception of Ref. [7], these parameter estimation approaches provide single point estimates for the parameters. They require direct access to the model being calibrated since either forward sensitivities or associated adjoint coefficients need to be estimated in processes that require derivative information and might have large memory footprint. Finally, there might also be an expectation of calibration data being smooth and clean, a requirement that might not be met. An approach that takes a different tack is presented in Ref. [7], where inertial parameters of a delta robot are calibrated using a least squares estimator. Its limitation is that the parameters sought must appear linearly in the model, which happens to work for inertial parameters in multibody dynamics. Against this backdrop, the parameter calibration problem is approached herein in a Bayesian framework, a choice motivated by three observations: it works with noisy data; it is not intrusive, in that no access to the equations of the model or their sensitivities is required; and it produces confidence bounds for the results.

Bayesian inference can be used to calibrate unknown parameters $\theta$ in a computer model [8]. Compared to calibrating a model via classical optimization that identifies the optimal $\theta$ values, the Bayesian approach views the unknown parameters as random variables and aims at producing the associated conditional probability distributions, i.e., the posterior distribution $\pi(\theta) = p(\theta|y)$, given some data $y$. In other words, the approach produces information that speaks to how well a certain choice of model parameters $\theta$ explains the observed/experimental data $y$. Although Bayes' theorem provides a formula of the posterior distribution

$$p(\theta|y) = \frac{p(y|\theta)\,p(\theta)}{\int p(y|\theta)\,p(\theta)\mathrm{d}\theta} \qquad (1)$$

it is often difficult to directly evaluate that formula, particularly for the evidence term in the denominator when the posterior distribution is on a high-dimensional space. Therefore, Markov-chain Monte Carlo (MCMC) methods [9–11] have been widely used to sample from the posterior distribution. Then, one does not have a closed form for the posterior, but the samples obtained from MCMC can be used to approximately evaluate any expectation with respect to the posterior distribution, which is after all the task of practical interest.

---

[1]Corresponding author.
Manuscript received June 20, 2022; final manuscript received March 16, 2023; published online April 8, 2023. Assoc. Editor: Jozsef Kovecses.

The MCMC algorithm is setup to produce samples of the distribution. It is suitable for the task of Bayesian inference, as MCMC algorithm can be used when the distribution is only known up to a normalizing constant. This is done by trying to see whether a particular choice of parameters adequately explains the observed data, which calls for running a simulation with the candidate set of parameters. Sequential Monte Carlo (SMC) samplers [12,13] have been explored as an efficient alternative to classical MCMC. In general, SMC samples from a sequence of distributions $\{\pi_n(\theta)\}_{n=1}^N$, which could arise from the conditional distributions $\pi_n(\theta) = p(\theta|y_1,...,y_n)$ with a sequence of data $\{y_n\}_{n=1}^N$, or from a sequence of artificial intermediate distributions [14] that starts from an easy-to-sample distribution, e.g., the prior distribution, and gradually converges to the posterior distribution. Compared to MCMC, SMC is easier to parallelize and less likely to be trapped in local modes.

This contribution is organized as follows. Section 2 covers the fundamentals of the Bayesian inference process used. It discusses the Bayesian framework; the SMC sampler that helps one draw samples from the posterior distribution; and the probabilistic programing framework PyMC [15] used to code in Python the described inference framework. In Sec. 3, the Bayesian machinery is used to answer this question: for simulating vehicle dynamics over short time intervals, can a high-fidelity high-mobility multipurpose wheeled vehicle (HMMWV) Chrono model [16,17] be replaced by a simple model, with eight degrees-of-freedom, that runs faster given that its evolution is captured by a small set of differential equations? A discussion of the Bayesian calibration approach embraced makes up Sec. 4, in which we highlight limitations of the methodology and our use of it. Directions of future work and a series of conclusion wrap up the paper.

## 2 Methodology

Herein, the general problem setting is cast as

$$y = \mathcal{G}(\theta) + \varepsilon, \quad \varepsilon \sim N(0, \Gamma) \tag{2}$$

where $\mathcal{G}: \mathcal{X} \mapsto \mathcal{Y}$ denotes a computer model with unknown parameters $\theta \in \mathcal{X}$; $y \in \mathcal{Y}$ represents the available data with noise $\varepsilon$ that follows a zero-mean normal distribution with a covariance matrix $\Gamma$; and $\mathcal{X}$ and $\mathcal{Y}$ denote complete normed vector spaces. The goal is to estimate the posterior distribution $p(\theta|y)$ via Bayesian inference

$$p(\theta|y) \propto \exp\left(-\frac{1}{2}\|y - \mathcal{G}(\theta)\|_\Gamma^2\right) p(\theta) \tag{3}$$

where $p(\theta)$ is the prior distribution that carries some existing knowledge about the unknown parameters $\theta$. We use SMC to draw samples from the posterior distribution $p(\theta|y)$.

**2.1 Sequential Monte Carlo.** The Sequential Monte Carlo sampler [12,13] aims at sampling a sequence of distributions. In the context of Bayesian model calibration, SMC samples a sequence of artificial intermediate distributions [14]

$$p_n(\theta|y) \propto \left[\exp\left(-\frac{1}{2}\|y - \mathcal{G}(\theta)\|_\Gamma^2\right)\right]^{\phi_n} p(\theta)$$

where $\{\phi_n\}$ denotes a sequence of hyperparameters that starts from 0 and eventually converges to 1. In practice, SMC employs sequential importance sampling to sample each $p_n$ based on an importance distribution $\eta_n(\theta)$. More specifically, it often starts with $\eta_0(\theta) = p(\theta)$ and then sequentially updates $\eta_n$ according to $\eta_{n-1}$ and the local random walk kernel $K_n(\theta, \theta')$ for $n \geq 1$ [13]. Compared to the traditional MCMC approach, SMC has a few advantages, including better sampling of multimodal distributions and more straightforward parallel implementation that improves the computational efficiency.

**2.2 Software Framework.** The workflow for Bayesian calibration of the vehicle model is presented in Fig. 1. Within this workflow, a Python package called PyMC [15] is used to sample from the posterior distribution.

*2.2.1 Preproccessing.* Due to the lack of experimental data, we use Chrono and its Chrono::Vehicle module [17] to generate synthetic high-fidelity "experimental" data. Since real data collected in physical experiments is noisy, we add Gaussian white noise on top of the data produced by the simulator. More details about the data generation process are provided in Sec. 3.1. We also define an 8DOF model whose unknown parameters $\theta$ we aim to calibrate are based on the synthetic data. The goal is to identify a choice of 8DOF model parameters that makes its dynamics a good proxy for that of the "real" vehicle in Chrono.

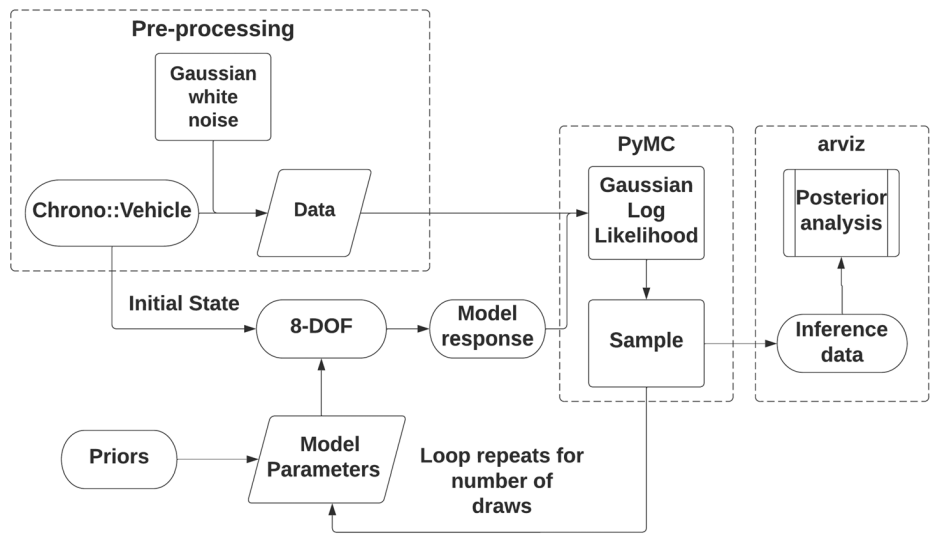

**Fig. 1 Workflow for Bayesian calibration of the vehicle model. Data synthetically generated in Chrono is used to setup the 8DOF model whose dynamics is a good proxy for that of the actual vehicle. The approach requires a Gaussian log likelihood function and a sampler, which we chose from PyMC. Once sampling is completed, PyMC provides the posterior as an inference data object which is used with ArviZ to perform analysis. See Ref. [19] for a Python implementation of this workflow.**

*2.2.2 Sampling.* We use a Gaussian likelihood function, i.e., the squared exponential term in Eq. (3). The likelihood function demands the evaluation of $\mathcal{G}(\theta)$, which involves running the 8DOF model and gathering model responses. For the prior distributions of the unknown parameters $\theta$, we assume a uniform distribution within an empirically chosen range for each unknown parameter. The samples obtained from PyMC, along with various sampler statistics, can be used with ArviZ [18] to do exploratory analysis of PyMC results.

## 3 Demonstration of Bayesian Calibration Framework

In robotics, one needs an expeditious model for tasks such as state estimation, path planning, or model predictive control. While typically task specific, these models are similar in one regard: they are expected to run much faster than real-time. These observations motivate the case study in this section: producing a simple vehicle model that runs fast and whose time evolution (dynamics) is a good proxy for the dynamics of a complex vehicle of interest. The vehicle of interest is simulated in Chrono to produce synthetic training/ground truth data and is subsequently used to calibrate a simplified vehicle model. Upon completing the calibration process, we assess how close the dynamics of the simple model comes to that of the vehicle of interest.

In this contribution, the simple model is an 8DOF model, while the vehicle of interest is a U.S. Army HMMWV. The Chrono high fidelity model used for simulating this HMMWV captures all relevant dynamics for a wide range of on- and off-road mobility applications. Constructed using the template-based approach enabled by Chrono::Vehicle [17], this model includes accurate models of the HMMWV independent double A-arm suspensions, a Pitman-arm or rack-and-pinion steering mechanism, a full dynamic model of the driveline (including models of the central and axle differentials), as well as models of the engine (based on power-torque curves including engine losses) and of the automatic transmission (torque converter and gear box). The main components of the resulting multibody system, which includes both kinematic joints and bushings, are shown in the simulation snapshots of Fig. 2. Chrono::Vehicle offers a variety of different tire models, from the commonly used semi-empirical Pacejka, Fiala, and TMeasy, to full FEA models using ANCF or corotational elements.

The 8DOF vehicle dynamics model is shown in Fig. 3. It has lateral, longitudinal, yaw, and roll DOFs. Additionally, four more differential equations are associated with the rotation of the four wheels [20–22]. The limitation of this model is that it does not capture the pitch and heave motions, and the front and rear suspensions are represented simply by their respective equivalent roll stiffness ($k_{\phi f}/k_{\phi r}$) and roll damping coefficients ($b_{\phi f}/b_{\phi r}$), thus it generally cannot be used to model aggressive acceleration/deceleration maneuvers. The equations of motion for the chassis velocities are thus given by [20,22]

$$m_t(\dot{u} - \omega_z v) = \sum F_{xgij} + (m_{uf}a - m_{ur}b)\omega_z^2 - 2h_{rc}m\omega_z\omega_x \quad (4)$$

$$m_t(\dot{v} + \omega_z u) = \sum F_{ygij} + (m_{ur}b - m_{uf}a)\dot{\omega}_z + h_{rc}m\dot{\omega}_x \quad (5)$$

$$\begin{aligned}J_z\dot{\omega}_z + J_{xz}\dot{\omega}_x &= (F_{yglf} + F_{ygrf})a - (F_{yglr} + F_{ygrr})b \\ &+ \frac{(F_{xgrf} - F_{xglf})c_f}{2} + \frac{(F_{xgrr} - F_{xglr})c_r}{2} \\ &+ (m_{ur}b - m_{uf}a)(\dot{v} + \omega_z u)\end{aligned} \quad (6)$$

$$\begin{aligned}\hat{J}_x\dot{\omega}_x + J_{xz}\dot{\omega}_z &= mgh_{rc}\phi - (k_{\phi f} + k_{\phi r})\phi - (b_{\phi f} + b_{\phi r})\dot{\phi} \\ &+ h_{rc}m(\dot{v} + \omega_z u)\end{aligned} \quad (7)$$

where $h_{rc} \equiv \frac{h_{rcf}b + h_{rcr}a}{a+b}$ and $\hat{J}_x \equiv (J_x + mh_{rc}^2)$.

In these equations, $h_{rcf}$ and $h_{rcr}$ are the vertical distances of the front and rear roll centers below the sprung mass center of mass (CM), and thus $h_{rc}$ is the vertical distance from the sprung mass CM to the vehicle roll center. It should be noted that Eq. (7) for the roll degree-of-freedom is written by considering moments acting about the vehicle roll center rather than the sprung mass CM, and thus the roll inertia of the sprung mass about the vehicle roll center ($J_x + mh_{rc}^2$) is considered in Eq. (7) [20,22]. The forces $F_{xgij}$ and $F_{ygij}$ are the global longitudinal and lateral forces at the four tire contact patches and the subscript "ij" denotes lf (left-front), rf (right-front), lr (left-rear), and rr (right-rear).

Using the notation in Ref. [23], a steady-state implementation of the Fiala tire model has been used in the development of these tire forces. The longitudinal and lateral slips used in the tire model were thus calculated respectively as

$$s_{ij} = \frac{r_{ij}\omega_{ij} - [u_{gij}\cos(\delta) + v_{gij}\sin(\delta)]}{|u_{gij}\cos(\delta) + v_{gij}\sin(\delta)|} \quad (8)$$

$$\alpha_{ij} = \tan^{-1}\left(\frac{v_{gij}}{u_{gij}}\right) - \delta \quad (9)$$

where $\delta$ is the steering angle at the tire (only the front wheels can be steered), $u_{gij}$ and $v_{gij}$ are the global longitudinal and lateral velocities of the tire contact patch, $\omega_{ij}$ is the angular velocity of the tire, and $r_{ij}$ is the instantaneous tire radius. For instance, the velocities of the tire contact patch at the right front contact patch are given by $u_{grf} = u + \omega_z c_f/2$ and $v_{grf} = v + \omega_z a$.

The angular velocity of the tire $\omega$ is given by the tire rotational dynamics model as [24] $J_w\dot{\omega}_{ij} = T_{dij} - T_{bij} - T_{rij} - r_{ij}F_{xtij}$, where $T_{dij}$ is the driving torque applied to the wheel, $T_{bij}$ is the braking torque applied to the wheel, $T_{rij}$ is the rolling resistance torque acting on the tire, $J_w$ is the rotational inertia of the wheel about its local Y axis (ISO-W coordinate system) and $F_{xtij}$ is the local longitudinal force on the tire contact patch. The drive torque and the brake torque are controls that are provided as input to the 8DOF model whereas the rolling resistance torque is given by $T_{rij} = -rr|F_z|\text{sgn}(\omega)$, where rr is the tire rolling resistance coefficient. The instantaneous tire radius r is calculated as [20] $r_{ij} = r0_{ij} - x_{tij}$, where $r0_{ij}$ is the unloaded radius of the tire and $x_{tij}$ is the instantaneous tire compression. Since the wheels do not have a DOF along the Z axis, $x_{tij}$ is calculated by a linear equation given by $x_{tij} = \frac{F_{zij}}{k_{tij}}$, where $F_{zij}$ is the vertical force on the tire and $k_{tj}$ is the front/rear tire vertical stiffness. Thus, it is assumed that the tire has no vertical damping properties and behaves as a linear spring on application of a vertical force. This vertical force $F_{zij}$ is obtained based on quasi-static lateral and longitudinal load transfers as [20,24]

$$F_{zglf} = \frac{mgb}{2(a+b)} + \frac{m_{uf}g}{2} - f_f - \frac{(k_{\phi f}\phi + b_{\phi f}\dot{\phi})}{c_f} - f_c \quad (10)$$

$$F_{zgrf} = \frac{mgb}{2(a+b)} + \frac{m_{uf}g}{2} + f_f + \frac{(k_{\phi f}\phi + b_{\phi f}\dot{\phi})}{c_f} - f_c \quad (11)$$

$$F_{zglr} = \frac{mga}{2(a+b)} + \frac{m_{ur}g}{2} - f_r - \frac{(k_{\phi r}\phi + b_{\phi r}\dot{\phi})}{c_r} + f_c \quad (12)$$

$$F_{zgrr} = \frac{mga}{2(a+b)} + \frac{m_{ur}g}{2} + f_r + \frac{(k_{\phi r}\phi + b_{\phi r}\dot{\phi})}{c_r} + f_c, \quad (13)$$

where $f_f \equiv \frac{m_{uf}h_{uf}}{c_f} + \frac{mb(h_{cg}-h_{rcf})}{c_f(a+b)}(\dot{v} + \omega_z u)$, and $f_r \equiv (\frac{m_{ur}h_{ur}}{c_r} + \frac{ma(h_{cg}-h_{rcr})}{c_r(a+b)})(\dot{v} + \omega_z u)$, and $f_c \equiv \frac{(mh_{cg} + m_{uf}h_{uf} + m_{ur}h_{ur})(\dot{u}-\omega_z v)}{2(a+b)}$.

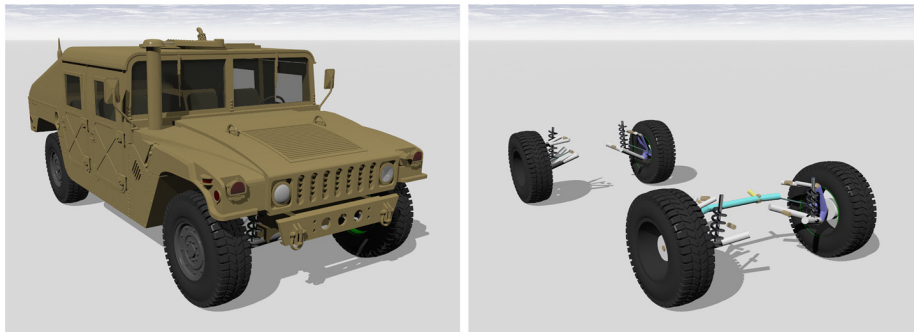

**Fig. 2 Chrono::Vehicle HMMWV multibody system model. The powertrain and driveline models are not rendered.**

Finally, using Eqs. (8) and (9), with the comprehensive slip $s\alpha$ and the coefficient of friction $U$ as defined in conjunction with the tire vertical forces $F_z$, the critical longitudinal slip ratio $s_{\text{critical}}$ is calculated as in Ref. [23]. Specifically, if the absolute value of the longitudinal slip ratio $|s| \leq s_{\text{critical}}$, then the tire is in elastic deformation state and the local longitudinal tire force is given by $F_{xtij} = C_{xj}s_{ij}$. Otherwise, the tire is in the complete sliding state and the local longitudinal tire force is $F_{xtij} = \text{sgn}(s)(F_{x1ij} - F_{x2ij})$, where $F_{x1ij} = U_{ij}|F_{zij}|$, $F_{x2ij} = |\frac{(U_{ij}F_{zij})^2}{4s_{ij}C_{xj}}|$, and $C_{xj}$ is the longitudinal tire stiffness of the front/rear wheels.

When the tire is in the elastic deformation state, the lateral tire force is given as (dropping tire index ij) $F_y = -U|F_z|(1-H^3)\text{sgn}(\alpha)$, where $H = 1 - \frac{C_y|\tan(\alpha)|}{3U|F_z|}$. In the complete sliding state, the lateral tire force is given as $F_y = -UF_z\text{sgn}(\alpha)$, where $C_y$ is the lateral tire stiffness, and $F_{xg}$ and $F_{yg}$ are obtained from $F_{xt}$ and $F_{yt}$ using a coordinate transformation.

The parameters of the 8DOF vehicle model are summarized in Table 1. Therein, a distinction is made between a tire and a wheel: the former is regarded as a force elements, the latter combines a tire and a wheel hub.

### 3.1 Data Generation.
The calibration of the 8DOF model using HMMWV synthetic data is split into three stages which concentrate, respectively, on the longitudinal dynamics, lateral dynamics, and tire rolling resistance. The longitudinal dynamics stage of the calibration process uses the vehicle longitudinal velocity along with the front and rear angular velocities of the wheels. The lateral dynamics stage uses the lateral velocity of the vehicle, as well as its yaw rate, roll angle, and roll rate. The tire rolling resistance stage of the calibration uses the vehicle's longitudinal velocity, but with a different vehicle maneuver.

A Gaussian white noise is added to the synthetic data. Although this is not a prerequisite to using Bayesian calibration approaches, noise is added to demonstrate that the data used need not be smooth, which is the case when dealing with real world scenarios. The standard deviation of the noise for each data vector is $\sigma_u = 0.1$, $\sigma_{\omega_f} = 1$, $\sigma_{\omega_r} = 1$, $\sigma_v = 0.05$, $\sigma_{\dot\psi} = 0.02$, $\sigma_\phi = 0.005$ and, $\sigma_{\dot\phi} = 0.002$.

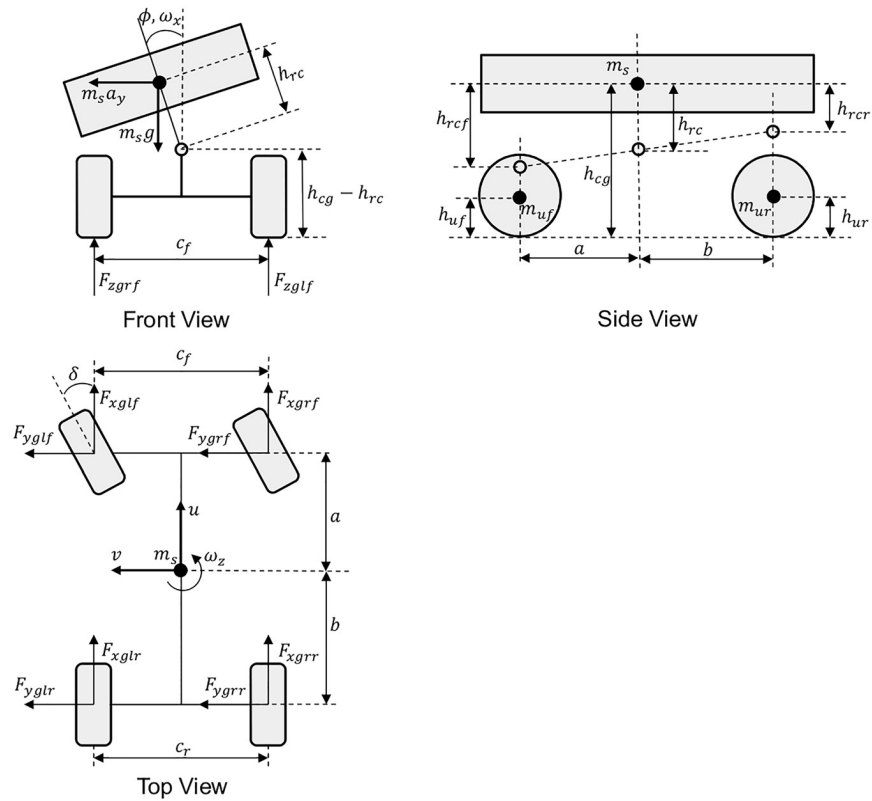

Fig. 3 Schematic of the 8DOF vehicle model [20,22]

**Table 1 Description of 8DOF model parameters**

| | Parameter description |
|---|---|
| $m$ | Sprung mass (kg) |
| $m_{uf}$ | Front unsprung mass (kg) |
| $m_{ur}$ | Rear unsprung mass (kg) |
| $J_x$ | Sprung mass roll inertia (kgm$^2$) |
| $J_z$ | Sprung mass yaw inertia (kgm$^2$) |
| $J_{xz}$ | Sprung mass XZ inertia (kgm$^2$) |
| $a$ | Distance of sprung mass CM from front axle (m) |
| $b$ | Distance of sprung mass CM from rear axle (m) |
| $h$ | Sprung mass CM height (m) |
| $c_f$ | Front track width (m) |
| $c_r$ | Rear track width (m) |
| $k_{\phi f}$ | Front roll stiffness (Nm/rad) |
| $k_{\phi r}$ | Rear roll stiffness (Nm/rad) |
| $b_{\phi f}$ | Front roll damping coefficient (Nm/s.rad) |
| $b_{\phi r}$ | Rear roll damping coefficient (Nm/s.rad) |
| $k_{tf}$ | Front tire stiffness (N/m) |
| $k_{tr}$ | Rear tire stiffness (N/m) |
| $C_{yf}$ | Front right tire lateral stiffness (N/rad) |
| $C_{yr}$ | Rear right tire lateral stiffness (N/rad) |
| $C_{xf}$ | Front right tire longitudinal stiffness (N) |
| $C_{xr}$ | Rear right tire longitudinal stiffness (N) |
| rr | Rolling resistance coefficient of tire (m) |
| $r_0$ | Nominal tire radius (m) |
| $J_w$ | Wheel roll inertia (kg m$^2$) |
| $h_{rcf}$ | Front roll center dist. below sprung mass CM (m) |
| $h_{rcr}$ | Rear roll center dist. below sprung mass CM (m) |

**3.2 Eight-Degrees-of-Freedom Known Parameters.** In order to calibrate the 8DOF model, we set out to first identify the parameters of interest that, in practice, are unknown or difficult to quantify using simple experiments. Measurable parameters such as the sprung/unsprung mass of the vehicle ($m, m_{uf}, m_{ur}$), the position of the CM ($h, a, b$) and the roll center ($h_{rcf}, h_{rcr}$), the track width ($c_f, c_r$) and the nominal tire radius ($r0$) are not required to be calibrated. Similarly, sprung mass inertias ($J_x, J_z, J_{xz}$), wheel inertias ($J_w$) and the tire vertical stiffness ($k_{tf}, k_{tr}$) are obtainable through fairly simple experiments. Thus, obtaining distributions over these parameters is not of much interest as they can be quantified fairly accurately without the need of data from vehicle maneuvers.

The parameters in Table 2 are considered known parameters, where the HMMWV model in the Chrono::Vehicle simulation and the 8DOF model use the same set of values. This leads to the following unknown parameters for which we obtain posterior distributions: $C_{xf}, C_{xr}, C_{yf}, C_{yr}, k_{\phi f}, k_{\phi r}, b_{\phi f}, b_{\phi r}$, rr.

**3.3 The Longitudinal Dynamics Phase of the Calibration.** We first calibrate the 8DOF model for longitudinal dynamics, i.e., we aim to make the 8DOF model perform like the HMMWV for longitudinal acceleration and deceleration vehicle maneuvers.

*3.3.1 Maneuver Description.* To generate calibration data, a simple straight line acceleration test is used. A normalized throttle input increasing from 0 to 0.5, see Fig. 4, is applied to the HMMWV vehicle. The HMMWV vehicle model uses a simple powertrain with a single gear ratio along with a maximum torque and speed map (much like a DC motor). The powertrain is connected to a four-wheel driveline which specifies the intermediate shaft inertias and the gear ratios of the differentials. The 8DOF model uses an identical powertrain and thus takes the same normalized throttle input. However, it does not have a driveline and so the torque from the powertrain is directly applied to the four wheels. It should be noted however that additional gear ratios of the differentials in the HMMWV vehicle are taken into account in the 8DOF model. Also, the 8DOF uses a simple tire model that has poor behavior when the vehicle starts from rest. Thus, the simulation of the 8DOF model starts from the state reached by the HMMWV at time 1.0 s into its simulation; at that point, the state of the 8DOF model (initial conditions) is made to match the state of the Chrono::Vehicle (red vertical line in Fig. 4). Although this does not play a role in the calibration effort, it remains a problem to be addressed in the future.

*3.3.2 Prior distribution.* Out of the unknown parameters, the ones identifiable through a longitudinal acceleration experiment are the front and rear longitudinal tire stiffness $C_{xf}$ and $C_{xr}$. According to the empirical knowledge about $C_{xf}$ and $C_{xr}$, we provide uniform-distributed prior distributions within a range, i.e., $C_{xf} \sim U(1000, 50000)$ and $C_{xr} \sim U(1000, 50000)$. The values 1000 and 50,000 are chosen to cover a reasonable range where we expect $C_{xf}$ and $C_{xr}$ to lie.

In addition to $C_{xf}$ and $C_{xr}$, we also sample the standard deviations of the noise in the three data vectors of interest: the longitudinal velocity of the vehicle, the left front wheel angular velocity and the left rear wheel angular velocity. Although in this study the standard deviation of the noise added is already known and can be directly used to evaluate the Gaussian likelihood, this may not be the case for real world data. Thus, it makes sense to sample the standard deviations together with other unknown model parameters as it can be particularly useful to build measurement error models when the noise is in fact unknown [25,26]. We choose a half-normal prior distribution as this incorporates our prior knowledge of the noise being small and always positive. The priors are of the form $\sigma_u \sim \mathcal{H}(0.1)$, $\sigma_{\omega_f} \sim \mathcal{H}(1)$, and $\sigma_{\omega_r} \sim \mathcal{H}(1)$.

**Table 2 Eight-degrees-of-freedom known parameters**

| | Parameter value |
|---|---|
| $M$ | 2097.85 (kg) |
| $m_{uf}$ | 127.86 (kg) |
| $m_{ur}$ | 129.98 (kg) |
| $J_x$ | 1289.00 (kgm$^2$) |
| $J_z$ | 4519.00 (kgm$^2$) |
| $J_{xz}$ | 3.26 (kgm$^2$) |
| $A$ | 1.68 (m) |
| $B$ | 1.68 (m) |
| $H$ | 0.71 (m) |
| $c_f$ | 1.82 (m) |
| $c_r$ | 1.82 (m) |
| $k_{tf}$ | 326332.00 (N/m) |
| $k_{tr}$ | 326332.00 (N/m) |
| $r_0$ | 0.47 (m) |
| $J_w$ | 11.00 (kgm$^2$) |
| $h_{rcf}$ | 0.38 (m) |
| $h_{rcr}$ | 0.32 (m) |

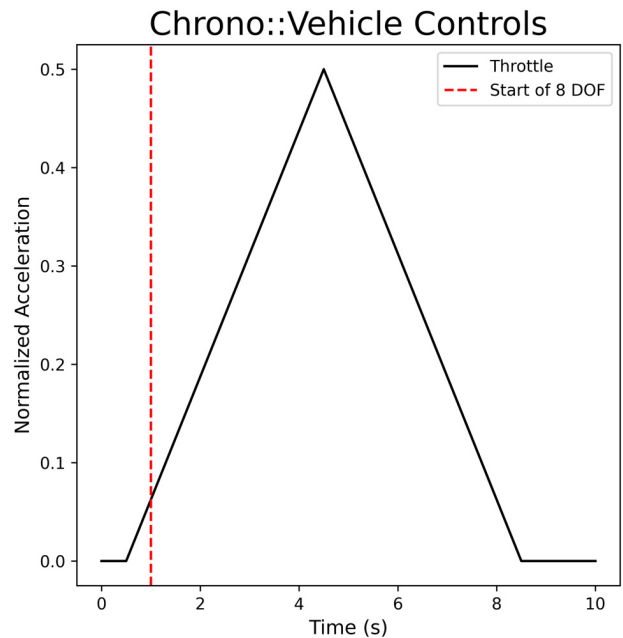

**Fig. 4** Normalized throttle input for longitudinal dynamics calibration

The vehicle maneuver for the longitudinal dynamics calibration is designed such that the remaining unknown variables $C_{yf}, C_{yr}, k_{\phi f}, k_{\phi r}, b_{\phi f}, b_{\phi r}$ play no effect on the response and can thus be randomly picked. The rolling resistance coefficient rr however does play an effect on the response, albeit a small one. Therefore, rr is not calibrated at this stage since its contribution to straight line acceleration performance is negligible. Instead, a simpler straight line deceleration vehicle maneuver is used in Sec. 3.5 to calibrate rr. We choose the values of the unknown parameters of interest as the mean of uniform prior distributions in Eq. (14). These prior distributions are further used in later calibration stage when we sample the remaining unknown parameters using additional data

$$C_{yf} \sim U(20000, 80000)$$
$$C_{yr} \sim U(20000, 80000)$$
$$k_{\phi f} \sim U(5000, 80000)$$
$$k_{\phi r} \sim U(5000, 80000) \quad (14)$$
$$b_{\phi f} \sim U(100, 30000)$$
$$b_{\phi r} \sim U(100, 30000)$$
$$rr \sim U(0.005, 0.03)$$

*3.3.3 Sampling.* The SMC sampler from PyMC is used for the longitudinal dynamics calibration. Eight SMC chains are run in parallel with each chain drawing 1000 samples. The SMC kernel used is Independent Metropolis Hastings. This means that in each chain, at each stage, the required (algorithm determined) steps of random walk (using Metropolis Hastings kernel) are run from each sample of SMC. The acceptance probability for each Metropolis Hastings chain is set to 0.9. Running multiple chains is essential for obtaining valid inferences from iterative simulations of finite length [27]. Having multiple chains is also critical in gauging chain convergence using diagnostics such as split-$\hat{R}$ [28]. In addition, multiple chains are more likely to reveal multimodality and poor adaptation or mixing [28]—scenarios that are particularly common for engineering applications. The minimum number of chains recommended in practice is four [28].

In practice, it is recommended that the Effective Sample Size (ESS) be greater than 400 [28]. It should be noted that ESS is not the same as the number of draws but rather a quantity of interest that captures how many independent draws contain the same amount of information as the dependent samples obtained by the Markov chains [28]. In this work, we have 1000 draws per SMC chain to ensure that ESS is greater than 400.

*3.3.4 Posterior.* The posterior distribution obtained for the parameters $C_{xf}$ and $C_{xr}$ is plotted using ArviZ in Fig. 5. The posterior distribution for the standard deviation of the synthetic Gaussian white noise added to the Chrono::Vehicle data is shown in Fig. 6. The posteriors obtained for the standard deviation of the noise are in the range of the values seen in Sec. 3.1. It is noted that the standard deviation obtained through the posterior is slightly larger than that supplied due to the difference in physics captured by the 8DOF and HMMWV models. Although these posteriors are useful as they show a distribution over the parameters, what is more insightful is to analyze the response of the 8DOF model with samples drawn from these posteriors—this will show one how the fact that we are not certain in the value of the calibration parameters reflects in the response (dynamics) of the 8DOF model.

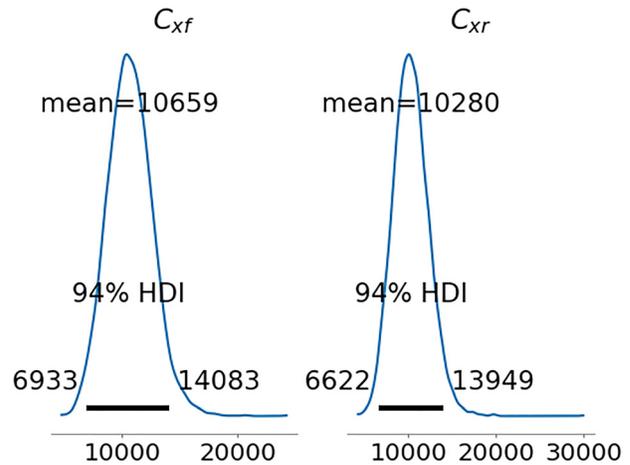

**Fig. 5 Longitudinal dynamics calibration—posterior distribution of $C_{xf}$ and $C_{xr}$. In the plot, HDI stands for high density interval (shown as a black horizontal line in the plot), which corresponds to the smallest interval that provides the required percentage value for a credible interval (in this case 94%).**

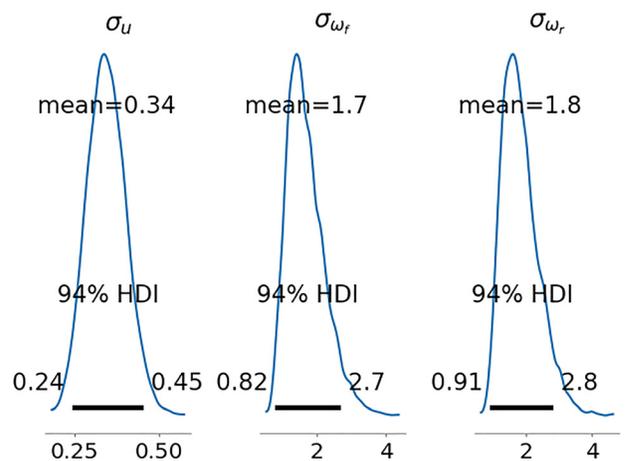

**Fig. 6 Longitudinal dynamics calibration—posterior distribution of $\sigma$**

*3.3.5 Eight-Degrees-of-Freedom Response.* We show the improvement of the posterior compared to the prior, by drawing samples of $C_{xf}$ and $C_{xr}$ from them, and assessing the fit between the 8-DOF response and that of the HMMWV. We first draw samples from the prior distribution, which represents our noninformative belief about the 8-DOF model prior to seeing any HMMWV data. Therefore, the 8DOF model can assume very different trajectories, see Figs. 7(a)–9(a). These figures are obtained by taking 100 random samples out of the prior distribution and plotting the 8DOF response. Bringing in the HMMWV data shapes our belief about the parameter values and leads to a posterior distribution. Note that drawing samples from the posterior is as straightforward as picking random samples from the Markov Chain that was used to generate the posterior. Thus, we pick 100 random samples of $C_{xf}$ and $C_{xr}$ from their joint posterior distribution obtained from the trace of the chain (to be discussed shortly) and plot 100 responses of the 8DOF model, see blue line in Figs. 7(b)–9(b).

These figures show that drawing samples from the posterior produces a response that is much narrower and closer to the noisy HMMWV data. Plotted along with the 100 responses are also the posterior/prior expectation and the posterior/prior mean. The posterior/prior expectation is obtained by taking the mean of the 100 responses whereas the posterior/prior mean is the response obtained by plugging in the mean values of the posterior/prior distributions of the calibrated parameters. The latter two lines are very close to each other, which suggests that the posterior mean is a good single point estimate of the posterior as it produces similar results to the expectation of the model response [29].

In order to quantify the posterior fit, we use the mean of the Root-Mean-Squared-Error (RMSE) between the 100 response lines and the noisy data. The RMSE of a single response with the data is given as

$$\text{RMSE} = \sqrt{\sum_{i=1}^{n_t} \frac{[\mathcal{G}(\theta)_i - y_i]^2}{n_t}}$$

where $\mathcal{G}(\theta)_i$ is the 8DOF model response at time-step $i$, $y_i$ is the noisy data at each time-step and $n_t$ is the number of time steps. The mean RMSE for both the prior and the posterior is reported in Table 3. This shows that the response of the 8DOF model with parameters drawn from the posterior produces a much tighter fit than with parameters drawn from the priors, a confirmation that the Bayesian framework uses the HMMWV data meaningfully when it comes to improving the quality of the 8DOF model.

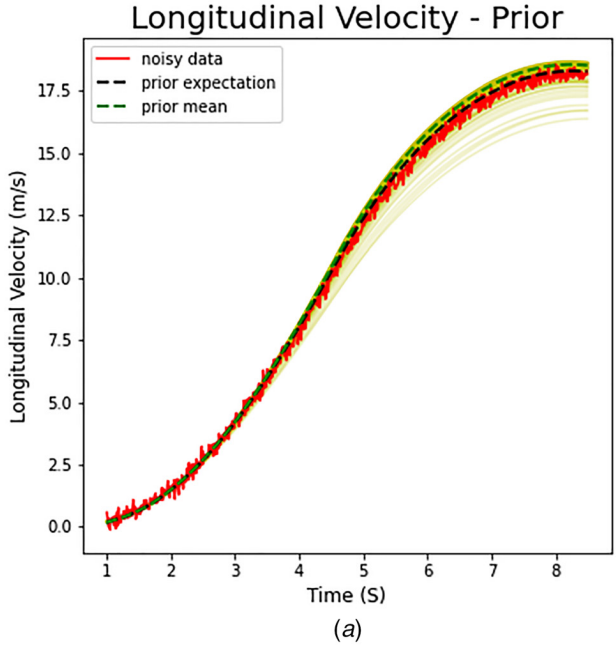

(a)

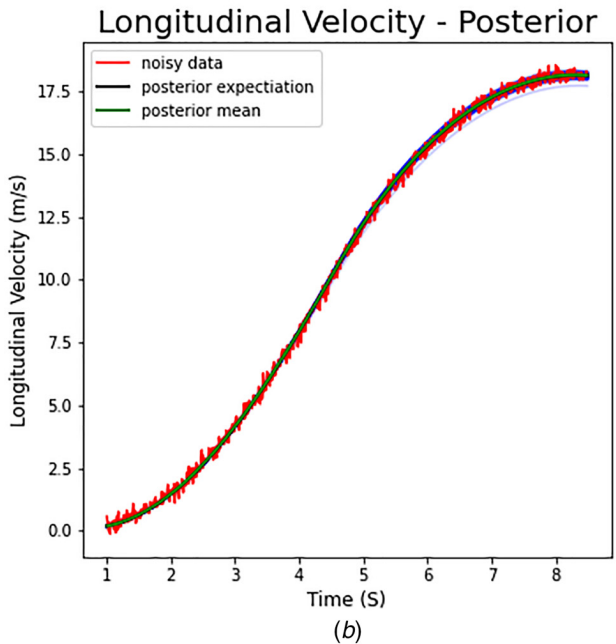

(b)

Fig. 7 Longitudinal dynamics calibration—comparison of the longitudinal velocity response of the 8DOF model with the noisy data with (a) parameters drawn from the prior (100 lines represent model responses for 100 draws of the parameters from their prior distribution) and (b) parameters drawn from the posterior (100 lines represent model responses for 100 draws of the parameters from their posterior distribution)

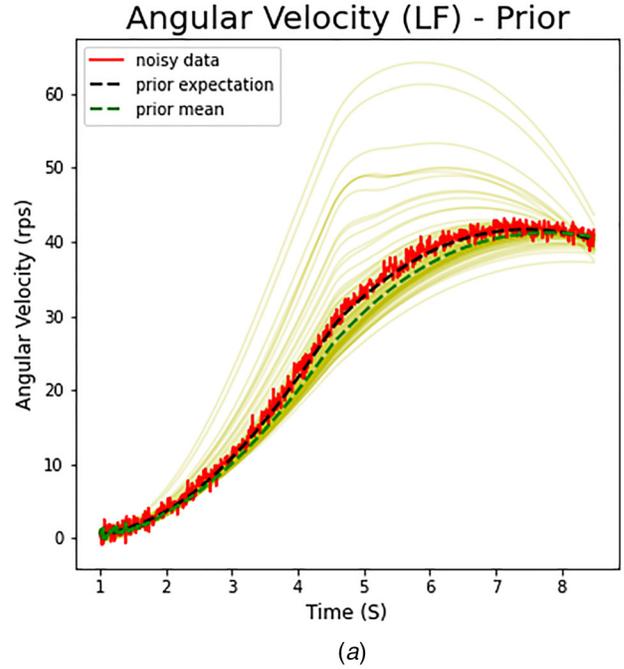

(a)

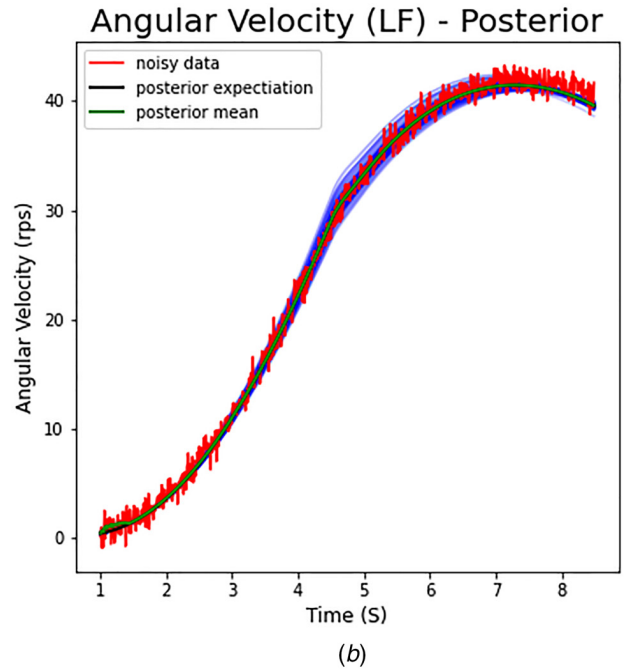

(b)

Fig. 8 Longitudinal dynamics calibration—comparison of the angular velocity (LF) response of the 8DOF model with the noisy data with (a) parameters drawn from the prior (100 lines represent model responses for 100 draws of the parameters from their prior distribution) and (b) parameters drawn from the posterior (100 lines represent model responses for 100 draws of the parameters from their posterior distribution)

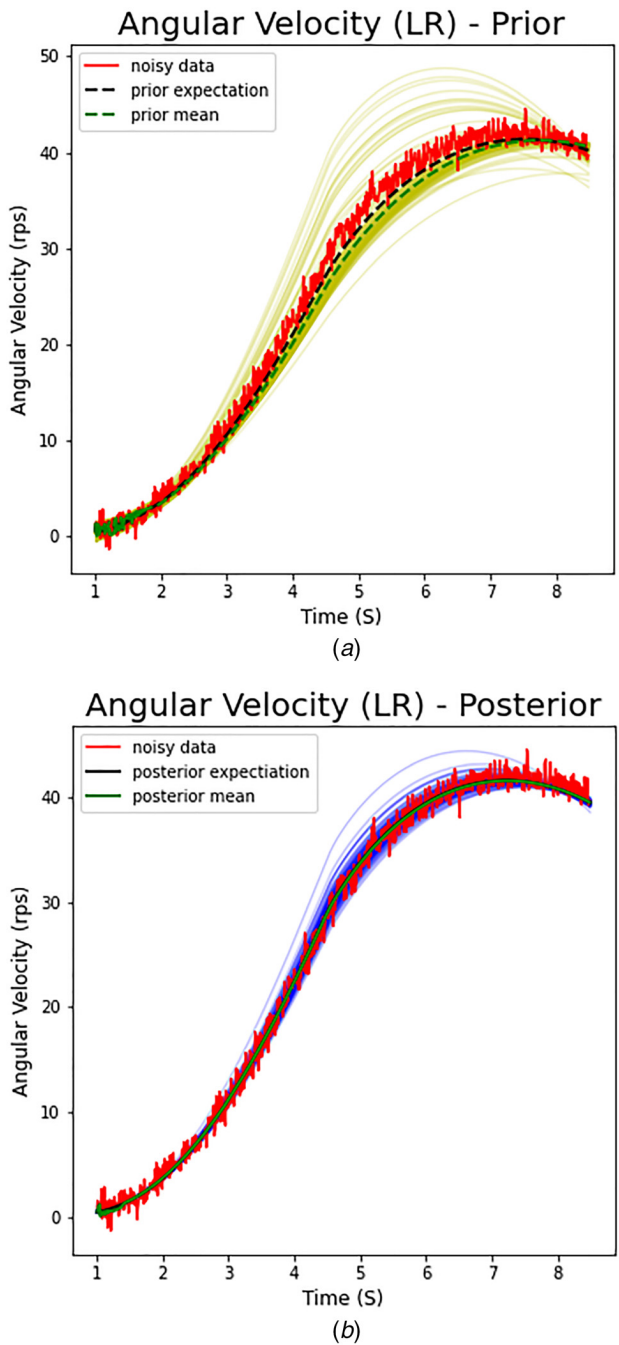

Fig. 9 Longitudinal dynamics calibration—comparison of the angular velocity (LR) response of the 8DOF model with the noisy data with (*a*) parameters drawn from the prior (100 lines represent model responses for 100 draws of the parameters from their prior distribution) and (*b*) parameters drawn from the posterior (100 lines represent model responses for 100 draws of the parameters from their posterior distribution)

Table 3  Mean RMSE—longitudinal dynamics calibration

| Data | Prior mean-RMSE | Posterior mean-RMSE |
|---|---|---|
| $U$ | 0.402 | 0.214 |
| $\omega_{lf}$ | 2.379 | 0.960 |
| $\omega_{rf}$ | 2.574 | 1.305 |

3.3.6 *Chain Diagnostics.* "While MCMC, as well as more general iterative simulation algorithms, can usually be proven to converge to the target distribution as the number of draws approaches infinity, there are rarely strong guarantees about their behavior after finite time" [28]. In order to draw conclusions from the posterior distribution, it is important to first assess how well the sampler is approximating the specified posterior distribution. One way to do this is by visually inspecting the sample paths of the eight chains via trace plots. The trace plots for the longitudinal dynamics calibration can be seen in Fig. 10 (right). Therein, it can be seen that the eight chains, which start from random points in prior parameter space, converge to a similar posterior. This shows that the chains are relatively insensitive to the starting point, thus showing geometrical ergodicity, a critical property for the central limit theorem to hold for approximate posterior expectations [28]. Generally, however, a qualitative assessment of chain convergence using visual inspection is often times considered "notoriously unreliable" [30]. Thus, we also calculate the split-$\hat{R}$ metric using ArviZ as this is the most commonly used convergence diagnostic in practice [28]. "Roughly speaking, the split $\hat{R}$ metric compares between and within chain estimates for model parameters" [28]. If the chain has not mixed well, the variance of all the chains mixed together will be higher than the variance of individual chains and the split-$\hat{R}$ will be larger than 1. It is recommended to only use samples if the split-$\hat{R} < 1.01$ [28]. Table 4 shows that the split-$\hat{R}$ for all our parameters is less than 1.01. In addition to this, we also calculate the mean and standard deviation of the Monte Carlo Standard Error (MCSE), which is a statistic that helps quantify the sampler error [28,31]. It can be seen from Table 4 that the mean and standard deviation of the MCSE for all our parameters calculated using ArviZ is much smaller than the scale of the parameters, which is reassuring.

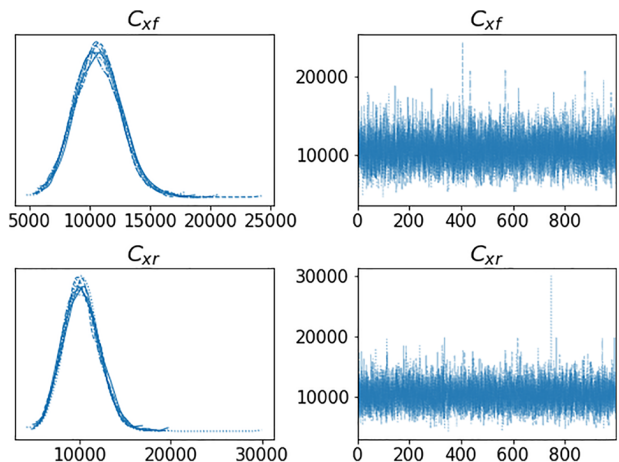

Fig. 10 Longitudinal dynamics calibration—(left) posterior of all eight chains (right) trace plots of all eight chains

Table 4  Chain statistics—$\hat{R}$ should be lesser than 1 and MCSE should be small

| Parameter | $\mu_{MCSE}$ | $\sigma_{MCSE}$ | $\hat{R}$ |
|---|---|---|---|
| $C_{xf}$ | 21.797 | 15.427 | 1.0003 |
| $C_{xr}$ | 22.687 | 16.043 | 1.0002 |
| $\sigma_u$ | 0.0006 | 0.0004 | 1.0001 |
| $\sigma_{\omega_f}$ | 0.0059 | 0.0042 | 1.0000 |
| $\sigma_{\omega_r}$ | 0.0060 | 0.0043 | 1.0002 |

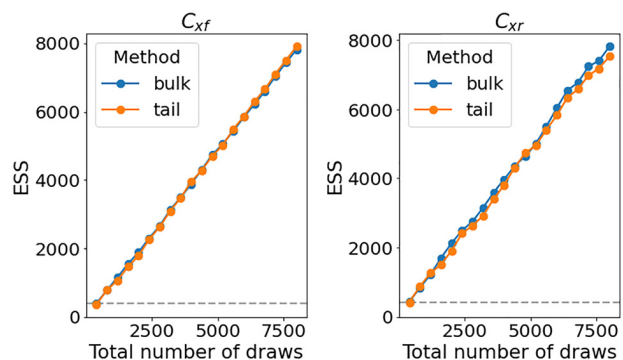

Fig. 11 Longitudinal dynamics calibration—bulk-ESS and tail-ESS evolution with the number of draws. For a well explored distribution, bulk/tail-ESS should increase linearly with the number of draws.

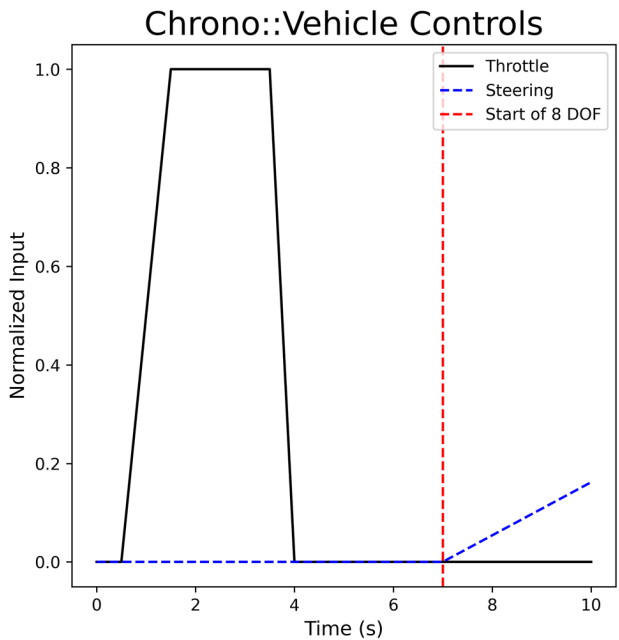

**Fig. 12 Normalized throttle and steering input for lateral dynamics calibration. The 8DOF model is set to the state of the Chrono:Vehicle at the red dotted line and given the same input controls.**

Further, we also plot the ESS evolution plot [28] as it gives a "scale-free" measure of information and is particularly useful when diagnosing sampling efficiency. In particular, we plot the evolution of both the bulk-ESS—the effective sample sizes in the "bulk" (5% to 95% quantiles) of the posterior, as well as the tail-ESS—the effective sample sizes in the "tails" of the posterior (minimum of 5% and 95% quantiles). "For a well explored distribution, we expect both the bulk-ESS and tail-ESS measures to grow linearly with the total number of draws S, or, equivalently, that the relative efficiency (ESS divided S) is approximately constant for different values of S" [28]. Figure 11 indicates that both the bulk-ESS and the tail-ESS for $C_{xf}$ and $C_{xr}$ increase linearly with the number of draws and reach a much higher value than the recommended value of 400 [28]. To conclude, the SMC sampler approximates the posterior distribution well.

### 3.4 Lateral Dynamics Calibration

*3.4.1 Maneuver Description.* To calibrate the parameters that dictate the lateral dynamics of the 8DOF model, the following Chrono HMMWV simulation is run to generate synthetic data: first, the vehicle is accelerated for 7 s to bring it to a speed of approximately 40 mph. For the next 3.7 s, the normalized steering input at the steering wheel is increased from 0 to 0.2, linearly (positive value corresponds to left steer). The throttle and steering input can be seen in Fig. 12. The HMMWV uses a rack-and-pinion steering mechanism. The 8DOF model is provided the state of the HMMWV at 7 s and is then given the same normalized steering input. Only data from 7 to 10.7 s is used for the calibration. This is done to separate the longitudinal dynamics from the lateral dynamics during the training stage. Unlike the HMMWV, the 8DOF model does not include a model of the steering mechanism. To ensure a similar steering angle at the front wheels for both the 8DOF and HMMWV for the same normalized steering input at the steering wheel, we take the average of the maximum front wheel steering angles of the HMMWV and set it as the maximum steering angles of the front wheels of the 8DOF model. A left/right average is used to accommodate Ackerman steering which is captured by the Chrono model, but not the 8DOF model. With this, the same normalized steering wheel angle is supplied to both the 8DOF model and HMMWV resulting in similar steering angles at the wheels.

*3.4.2 Prior Distribution.* The parameters calibrated during this stage of the process are $C_{yf}, C_{yr}, k_{\phi f}, k_{\phi r}, b_{\phi f}, b_{\phi r}$, along with the standard deviation of the noise of each of the data vectors of interest—$\sigma_v, \sigma_{\dot\psi}, \sigma_\phi, \sigma_{\dot\phi}$ (corresponding to lateral velocity, yaw angle, roll angle and roll rate). For the 8DOF model parameters, the same priors from Eq. (14) for $C_{yf}, C_{yr}, k_{\phi f}$ and $k_{\phi r}$ are used. However, for $b_{\phi f}$ and $b_{\phi r}$ we sample a single vehicle damping coefficient $b_\phi$ and then set $b_{\phi f} = b_{\phi r} = 0.5\, b_\phi$. This choice is further explained in Sec. 4.2. The mean of the posterior obtained in the longitudinal dynamics calibration for $C_{xf}$ and $C_{xr}$ is used. A half normal prior was chosen for the noise parameters. The scale of the prior is chosen according to the scale of the data as $\sigma_v \sim \mathcal{H}(0.05), \sigma_{\dot\psi} \sim \mathcal{H}(0.05), \sigma_\phi \sim \mathcal{H}(0.005)$, and $\sigma_{\dot\phi} \sim \mathcal{H}(0.005)$.

*3.4.3 Sampling.* The SMC sampler described for longitudinal dynamics is used herein; also, the SMC settings stay the same.

*3.4.4 Posterior.* The posterior distribution obtained for the parameters of interest is shown in Fig. 13. For the noise parameters, see Fig. 14. The posteriors are very narrow with a concentrated 94% HDI, which is highly desirable. It can also be seen that $b_{\phi f}$ and $b_{\phi r}$ have identical posteriors, which is expected based the way they are evaluated via $0.5 b_\phi$. Also, the standard deviation of the noise obtained is in the same range as those supplied during data generation (see Sec. 3.1). Similar to the longitudinal dynamics calibration, we will first discuss the 8DOF response using the posterior distribution and then discuss chain convergence statistics.

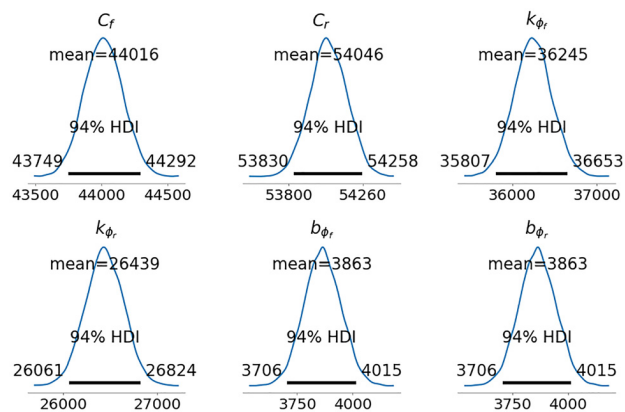

**Fig. 13 Posterior distribution of lateral parameters**

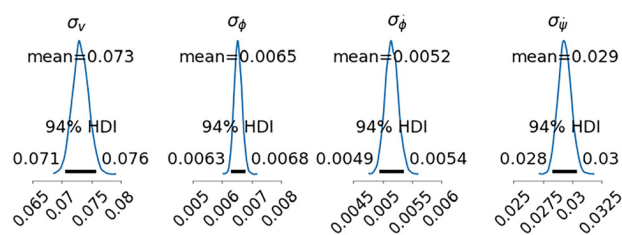

**Fig. 14 Posterior distribution of noise for lateral dynamics calibration**

*3.4.5 Eight-Degrees-of-Freedom Response.* A set of 100 responses using the prior distribution are summarized in Figs. 15(*a*)–18(*a*). We pick 100 random samples of $C_{yf}$, $C_{yr}$, $k_{\phi f}$, $k_{\phi r}$, $b_{\phi f}$ and $b_{\phi r}$ from the joint posterior distribution to plot 100 responses of the 8DOF model after bringing to bear the HMMWV synthetic data. The

Table 5  Mean RMSE—lateral dynamics calibration

| Data | Prior mean-RMSE | Posterior mean-RMSE |
|---|---|---|
| $V$ | 1.397 | 0.037 |
| $\phi$ | 0.025 | 0.003 |
| $\dot\phi$ | 0.02 | 0.002 |
| $\psi$ | 0.175 | 0.016 |

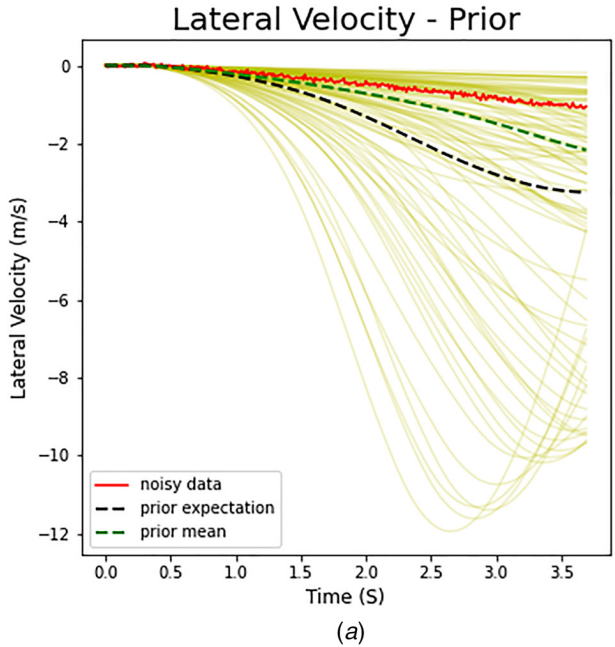

(*a*)

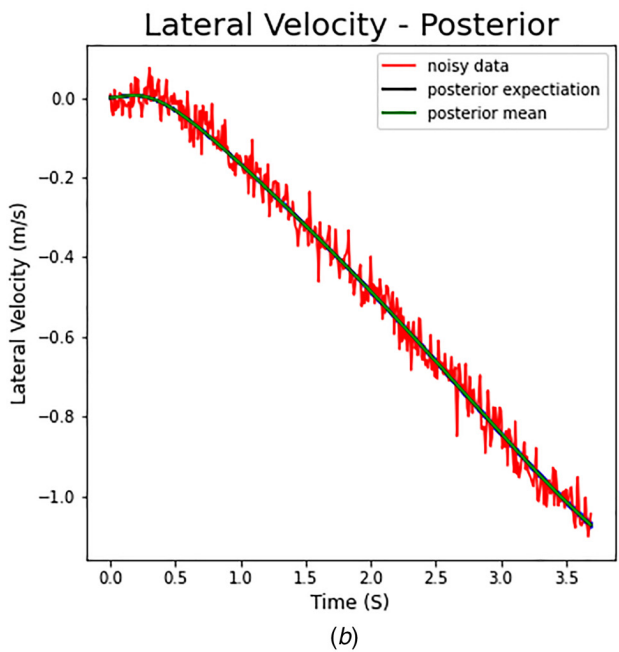

(*b*)

Fig. 15  Lateral dynamics calibration—comparison of the lateral velocity response of the 8DOF model with the noisy data with (*a*) parameters drawn from the prior (100 lines represent model responses for 100 draws of the parameters from their prior distribution) and (*b*) parameters drawn from the posterior (100 lines represent model responses for 100 draws of the parameters from their posterior distribution)

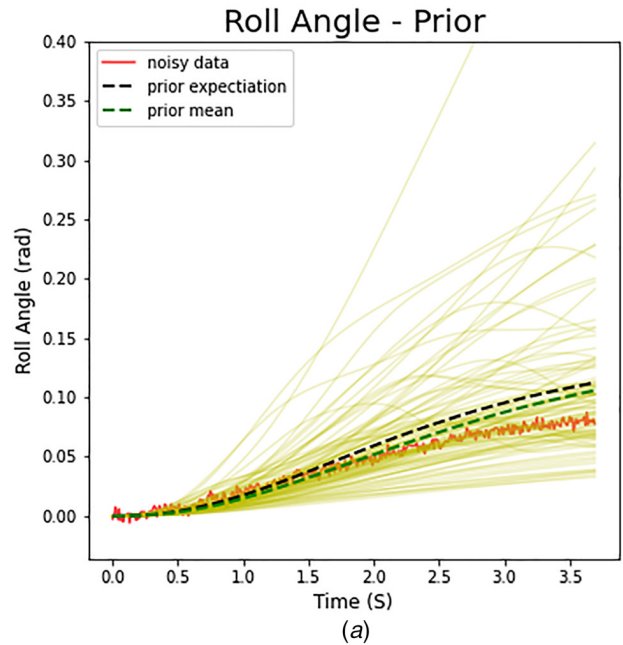

(*a*)

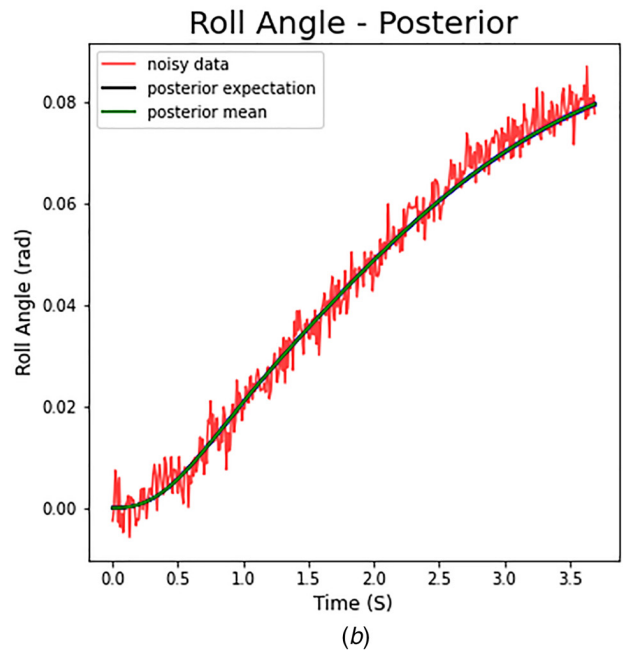

(*b*)

Fig. 16  Lateral dynamics calibration—comparison of the Roll Angle response of the 8DOF model with the noisy data with (*a*) parameters drawn from the prior (100 lines represent model responses for 100 draws of the parameters from their prior distribution) and (*b*) parameters drawn from the posterior (100 lines represent model responses for 100 draws of the parameters from their posterior distribution)

100 responses are seen in Figs. 15(*b*)–18(*b*). To quantify the posterior fit, we use again the mean of the RMSE between the 100 response lines and the noisy data. The results in Table 5 confirm that the average RMSE of the responses using samples from the posterior is approximately 10 times smaller than using samples from the prior across all data vectors.

*3.4.6 Chain Diagnostics.* An inspection of the trace plots for the eight chains in Fig. 19 reveals that the eight chains converge to a similar posterior. Table 6 shows that the split-$\hat{R}$ is less than 1.01 for all parameters. The mean and standard deviation of the MCSE is also small compared to the scale of each parameter and is thus acceptable. The bulk-ESS and tail-ESS evolution plots are shown in Fig. 20. The values obtained are well above the required minimum of 400 and also grow linearly with the number of draws.

The SMC sampler is thus able to approximate the posterior distribution well. The posteriors over the parameters $C_{yf}, C_{yr}, k_{\phi f}, k_{\phi r}, b_{\phi f}$ and $b_{\phi r}$ also produce a well calibrated lateral dynamics response. We now finally calibrate the rolling resistance of the tires using a similar approach. We then test our calibrated 8DOF model against different combinations of vehicle maneuvers that were not seen in the "training" data.

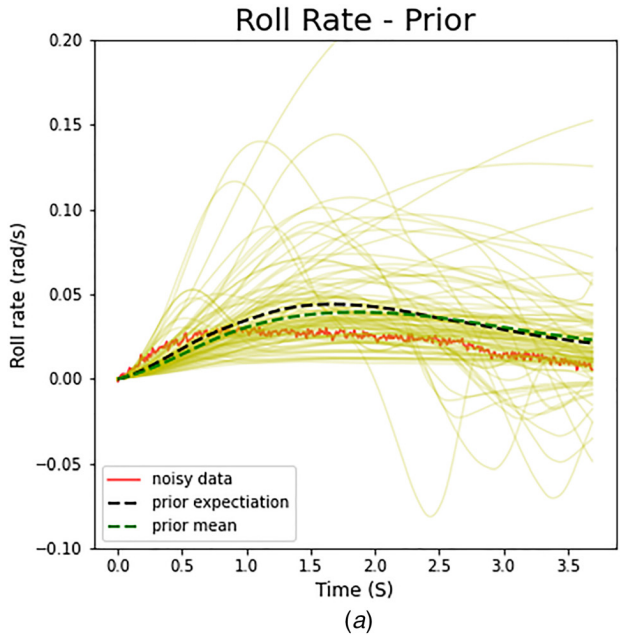

(*a*)

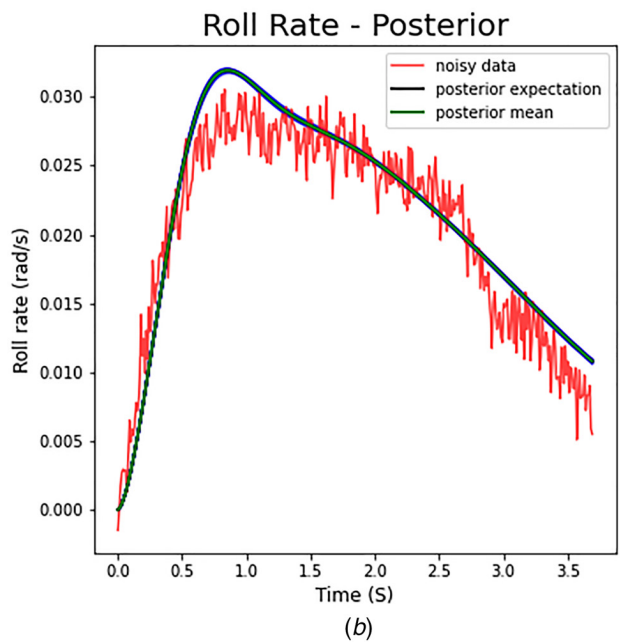

(*b*)

Fig. 17 Lateral dynamics calibration—comparison of the Roll rate response of the 8DOF model with the noisy data with (*a*) parameters drawn from the prior (100 lines represent model responses for 100 draws of the parameters from their prior distribution) and (*b*) parameters drawn from the posterior (100 lines represent model responses for 100 draws of the parameters from their posterior distribution)

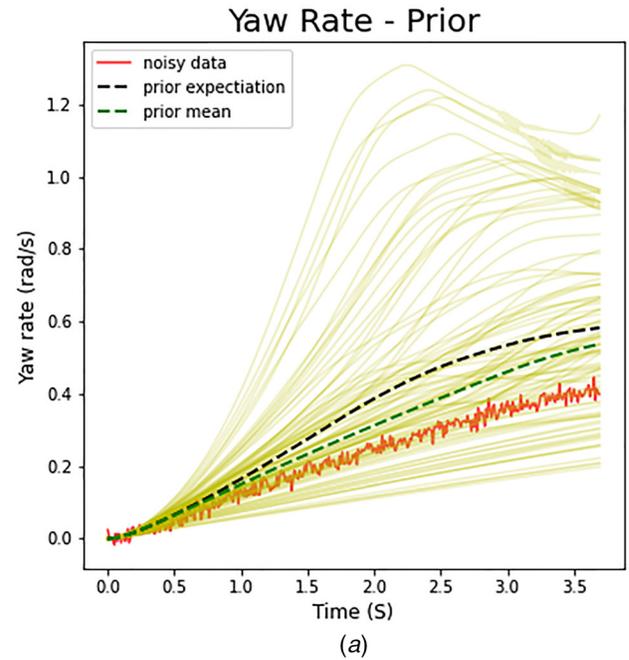

(*a*)

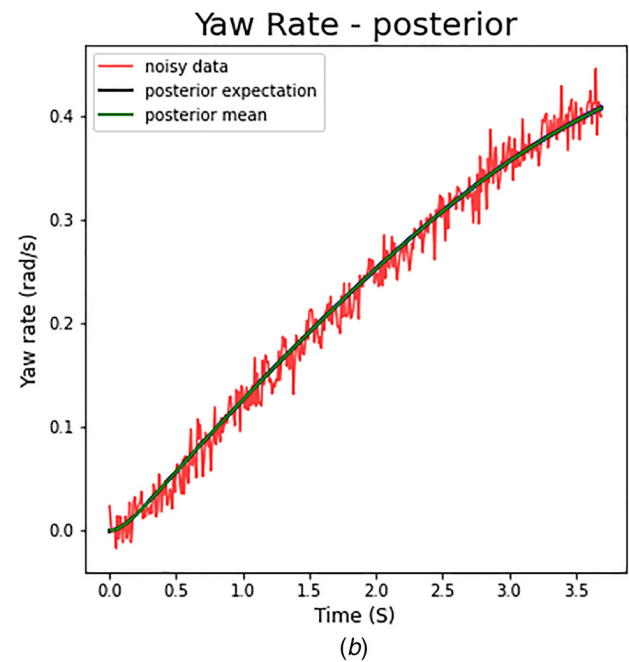

(*b*)

Fig. 18 Lateral dynamics calibration—comparison of the Yaw rate response of the 8DOF model with the noisy data with (*a*) parameters drawn from the prior (100 lines represent model responses for 100 draws of the parameters from their prior distribution) and (*b*) parameters drawn from the posterior (100 lines represent model responses for 100 draws of the parameters from their posterior distribution)

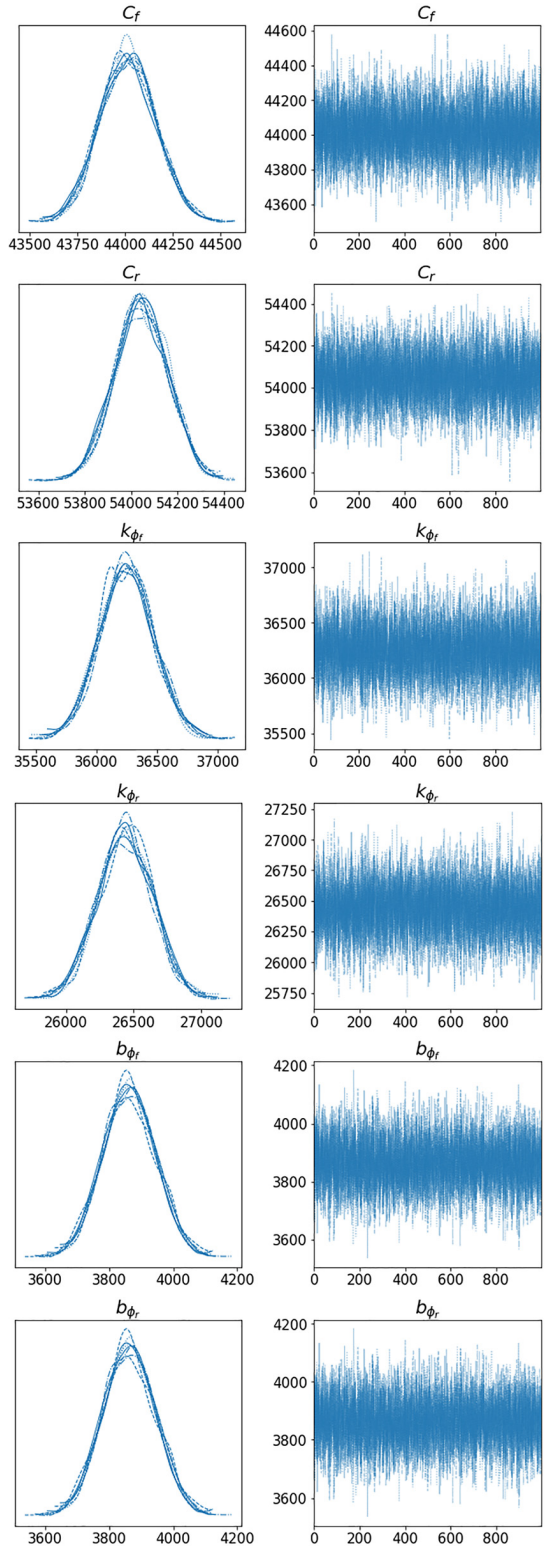

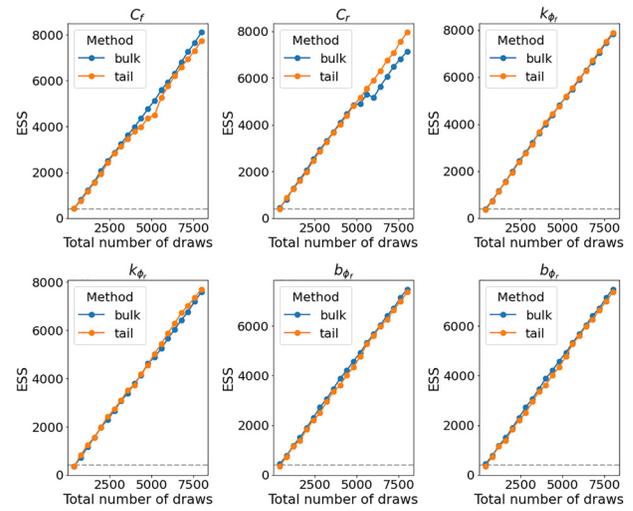

Fig. 20 Lateral dynamics calibration—bulk-ESS and tail-ESS evolution versus the number of draws. For a well explored distribution, bulk/tail-ESS should increase linearly with the number of draws.

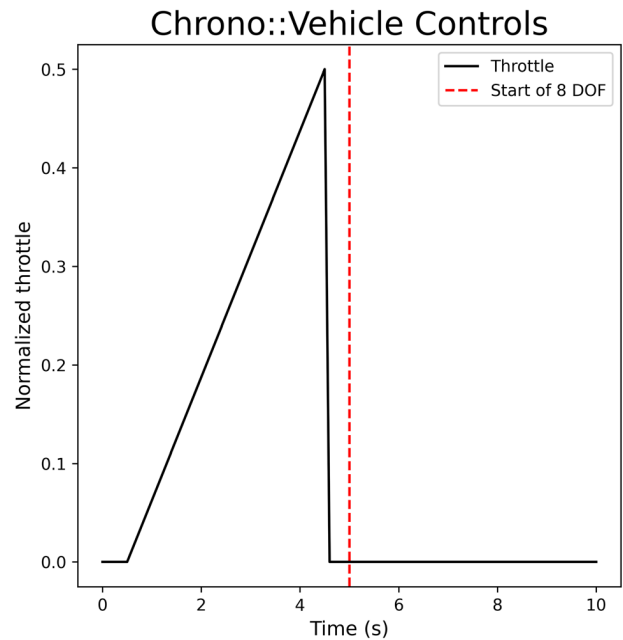

Fig. 21 Normalized throttle input for rolling resistance coefficient calibration. Note: The 8DOF model is set to the state of the Chrono: Vehicle at the dotted line and then given the same inputs.

Fig. 19 Lateral dynamics calibration—(left) posterior of all 8 chains (right) Trace plots of all eight chains

### 3.5 Rolling Resistance Calibration

*3.5.1 Maneuver Description.* This stage of the calibration process draws on a basic maneuver shown in Fig. 21. We first accelerate the HMMWV in Chrono to a speed of about 90 km/h. We then let the vehicle coast. Note that there is no aerodynamic drag included in the Chrono HMMWV model for the purpose of these simulations and therefore the slow down comes from the rolling resistance of the tires. The 8DOF model is set to the state of the HMMWV at the 5.6 s time-mark as shown by the dotted red line in Fig. 21. The Longitudinal velocity is selected as the data vector of interest that is used to drive the calibration process.

*3.5.2 Prior Distribution.* Since we expect the rolling resistance coefficient values of both the 8DOF tire model and the HMMWV tire model to be similar, we can provide a narrow prior around the HMMWV's value (Eq. (14)). We also provide a prior over the standard deviation of the longitudinal velocity noise as $\sigma_u \sim \mathcal{H}(0.1)$. This is the same prior for the noise used in the longitudinal dynamics component of the calibration effort.

*3.5.3 Sampling.* The SMC sampler described earlier, with the same settings, is re-used for the rolling resistance calibration.

*3.5.4 Posterior.* The posterior distribution obtained for the rolling resistance coefficient is shown in Fig. 22. For the posterior distribution of the standard deviation of the longitudinal velocity noise, see Fig. 23. The posteriors visually look narrow with a thin 95% high density interval (HDI).

Table 6  Chain statistics—$\hat{R}$ should be lesser than 1 and MCSE should be small

| Parameter | $\mu_{\text{MCSE}}$ | $\sigma_{\text{MCSE}}$ | $\hat{R}$ |
|---|---|---|---|
| $C_f$ | 1.615 | 1.142 | 1.0002 |
| $C_r$ | 1.358 | 0.961 | 1.0004 |
| $k_{\phi_f}$ | 2.542 | 1.798 | 1.0004 |
| $k_{\phi_r}$ | 2.345 | 1.659 | 1.0005 |
| $b_{\phi_f}$ | 0.959 | 0.0678 | 1.0007 |
| $\sigma_v$ | $1 \times 10^{-4}$ | $1 \times 10^{-4}$ | 1.0000 |
| $\sigma_\phi$ | $1 \times 10^{-5}$ | $1 \times 10^{-5}$ | 1.0000 |
| $\sigma_{\dot\phi}$ | $1 \times 10^{-5}$ | $1 \times 10^{-5}$ | 1.0002 |
| $\sigma_{\dot\psi}$ | $1 \times 10^{-4}$ | $1 \times 10^{-4}$ | 1.0002 |

Table 7  Mean RMSE—rolling resistance coefficient calibration

| Data | Prior mean-RMSE | Posterior mean-RMSE |
|---|---|---|
| $U$ | 0.354 | 0.133 |

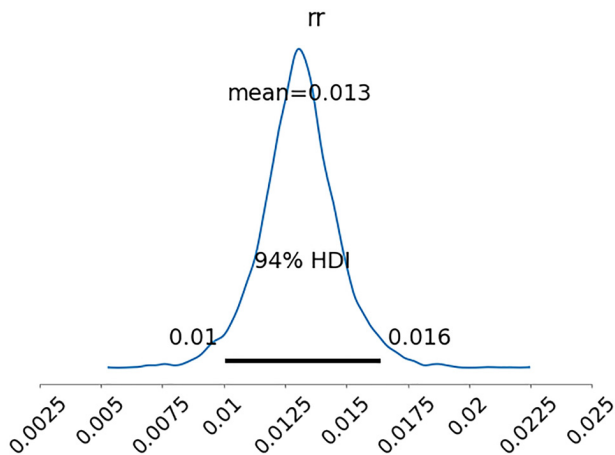

Fig. 22  Posterior distribution of the rolling resistance coefficient *rr*

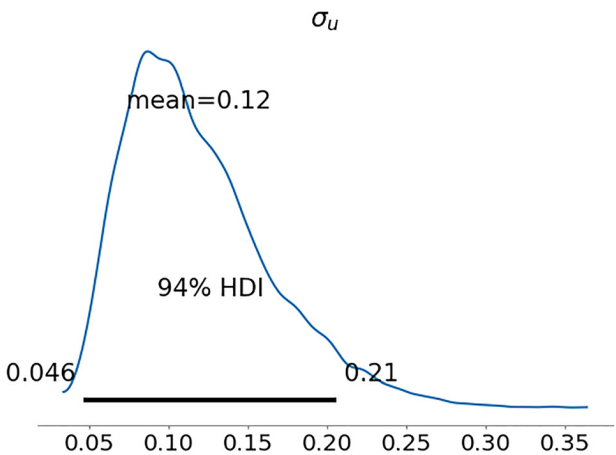

Fig. 23  Rolling resistance calibration—posterior distribution of $\sigma_u$

*3.5.5 Eight-Degrees-of-Freedom Response.* We perform a similar exercise to evaluate the response of the 8DOF postcalibration—draw 100 samples and plot the 8DOF system response, see Figs. 24(*a*) and 24(*b*). The RMSE results are provided in Table 7. Compared to the lateral calibration, the agreement between sampled simulations is less prominent, nonetheless the posterior mean and expectation show significant improvements over the prior.

*3.5.6 Chain Diagnostics.* We once again evaluate the trace (Fig. 25), the split-$\hat{R}$ (Table 8) and the ESS (Fig. 26). All the

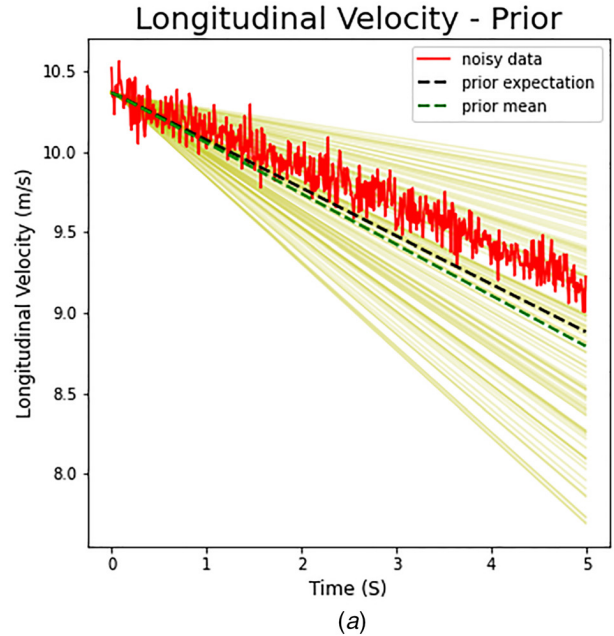

(*a*)

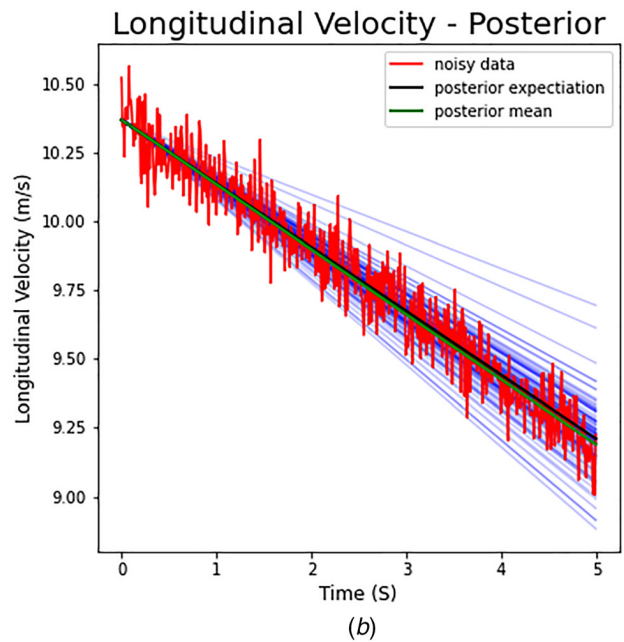

(*b*)

Fig. 24  Rolling resistance calibration—comparison of the longitudinal velocity response of the 8DOF model with the noisy data with (*a*) parameters drawn from the prior (100 lines represent model responses for 100 draws of the parameters from their prior distribution) and (*b*) parameters drawn from the posterior (100 lines represent model responses for 100 draws of the parameters from their posterior distribution

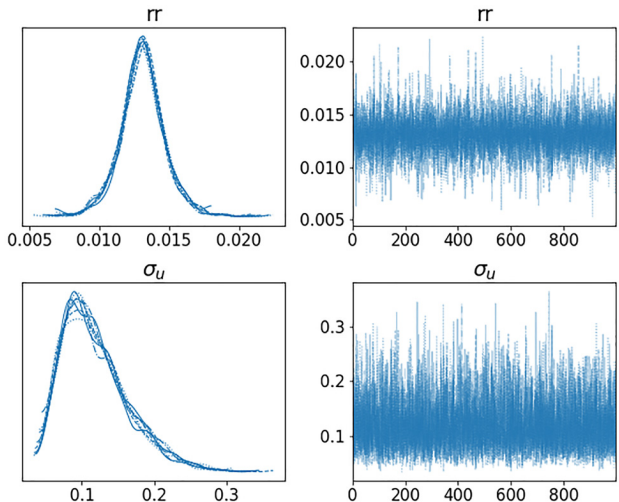

**Fig. 25** Rolling resistance calibration—(left) posterior of all eight chains (right) Trace plots of all eight chains

**Table 8** Chain statistics—$\hat{R}$ should be less than 1 and MCSE should be small

| Parameter | $\mu_{MCSE}$ | $\sigma_{MCSE}$ | $\hat{R}$ |
|---|---|---|---|
| rr | 0.000025 | 0.000018 | 0.9997 |
| $\sigma_u$ | 0.000664 | 0.000469 | 0.9998 |

convergence diagnostics suggest a successful calibration process, as the SMC sampler is able to approximate the posterior well.

### 3.6 Testing of the Calibrated Eight-Degrees-of-Freedom Model.
The calibration steps described in Subsections 3.3–3.5, which drew on data collected from the HMMWV over a small number of maneuvers, have lead to a calibrated 8DOF model. The question is whether the 8DOF model has become a good proxy for the HMMWV. To answer this, we subjected both the 8DOF model and HMMWV to a set of maneuvers that were not used during the calibration stage. We use the posteriors from the previous three calibration stages and plot 100 responses of the 8DOF model (Figs. 28–35). The control inputs chosen are illustrated in Fig. 27. We first accelerate the vehicle by applying a steeper ramp throttle from 0 to 3.5 s. We then let go of the throttle and apply a left step steer simultaneously. Then we apply a right ramp steer for approximately

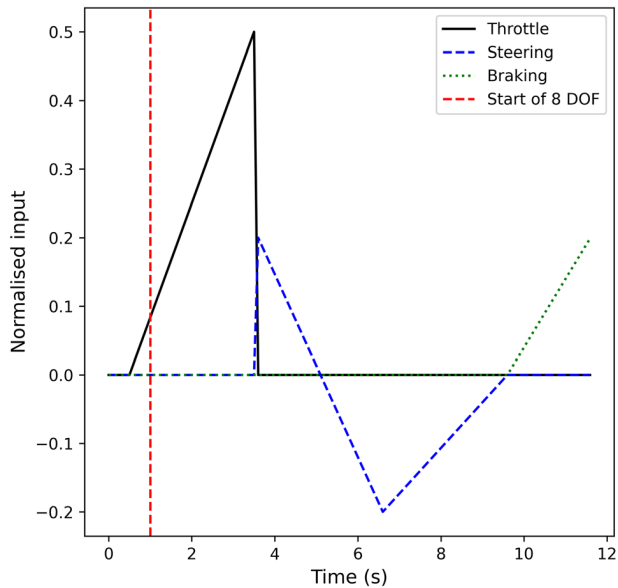

**Fig. 27** Normalized inputs for test maneuver. The 8DOF model is set to the state of the HMMWV at the red dotted line and then given the same inputs.

3 s. We then again bring back the steer to 0 and apply brakes for the last 2 s. We set the state of the 8DOF model as the state of the Chrono::Vehicle at the 1 s time-mark of the simulation. This is because, as mentioned before, the 8DOF has difficulties starting from rest.

Similarly we also calculate the mean-RMSE for each of the data vectors (Table 9). Overall, the responses of the 8DOF model is much more concentrated in and around the Chrono data postcalibration.

Figure 32 indicates that the 8DOF model roll angle during the first 3.5 s is relatively constant and close to zero. The evolution of the roll angle in the data is however decreasing. This is because the HMMWV has a small amount of roll even during straight-line acceleration due to the reaction torque of the engine and transmission on the chassis. This cannot be captured by the 8DOF model through any combination of parameters. Similarly, the 8DOF model is also not able to match the transient roll rate during the step steer maneuver (Fig. 33). The 8DOF roll rate does however stabilize to a value much closer to the data.

*3.6.1 Time Comparison.* The 11.6 s scenario described above will be used for a time comparison to gain a rough understanding of the 8DOF model execution speed. A time-step of $5 \times 10^{-3}$ s is used for the 8DOF model; the HMMWV simulation in Chrono uses a time-step of $2 \times 10^{-3}$ s and with an Euler linearized implicit integrator [16,32]. The 8DOF model uses a half-implicit integrator [33]. The simulations were performed on a *Intel(R) Core(TM) i7-7500 U CPU @ 2.70 GHz* and the results are provided in Table 10. Note that the HMMWV is simulated at a real-time factor (RTF) of about 0.41. Unsurprisingly, the 8DOF model runs more than an order of magnitude faster than the HMMWV, at an RTF of 0.03. The 8DOF simulation can be further accelerated by re-implementing the model in C/C++, a future task that falls outside the scope of this work.

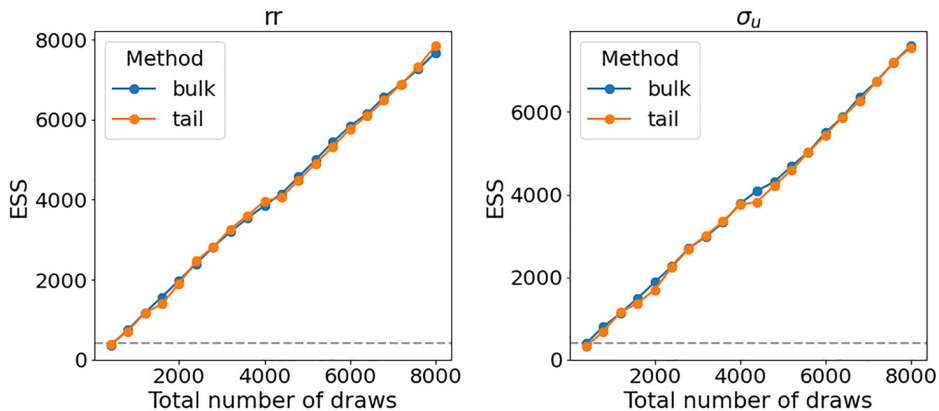

**Fig. 26** Rolling resistance calibration—bulk-ESS and tail-ESS evolution with the number of draws. For a well explored distribution, bulk/tail-ESS should increase linearly with the number of draws.

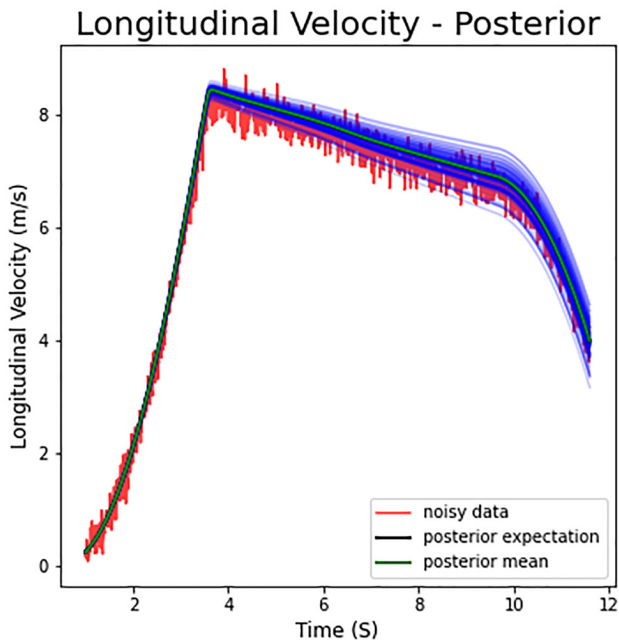

**Fig. 28** Comparison of the longitudinal velocity response of the 8DOF model with the noisy data with parameters drawn from the posterior

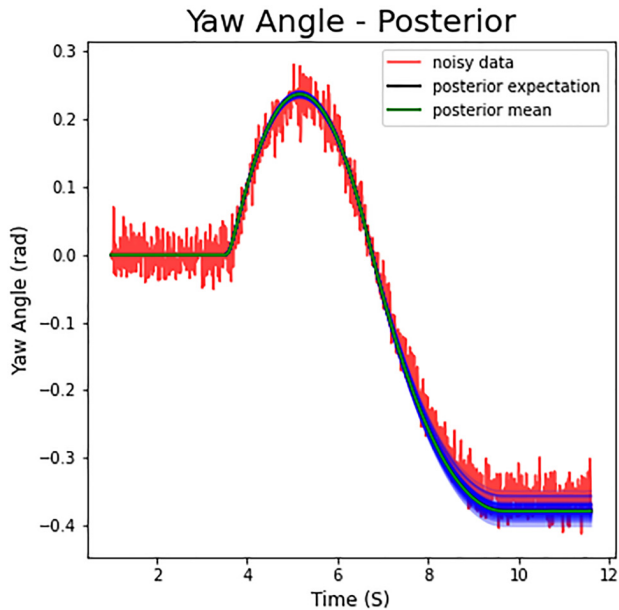

**Fig. 30** Comparison of the yaw angle response of the 8DOF model with the noisy data with parameters drawn from the posterior

**Table 9  Mean RMSE—test experiment**

| Data | Posterior mean-RMSE |
| --- | --- |
| $U$ | 0.319 |
| $V$ | 0.020 |
| $\psi$ | 0.026 |
| $\dot{\psi}$ | 0.021 |
| $\phi$ | 0.006 |
| $\dot{\phi}$ | 0.016 |
| $\sigma_{\omega_{lf}}$ | 1.300 |
| $\sigma_{\omega_{rr}}$ | 1.245 |

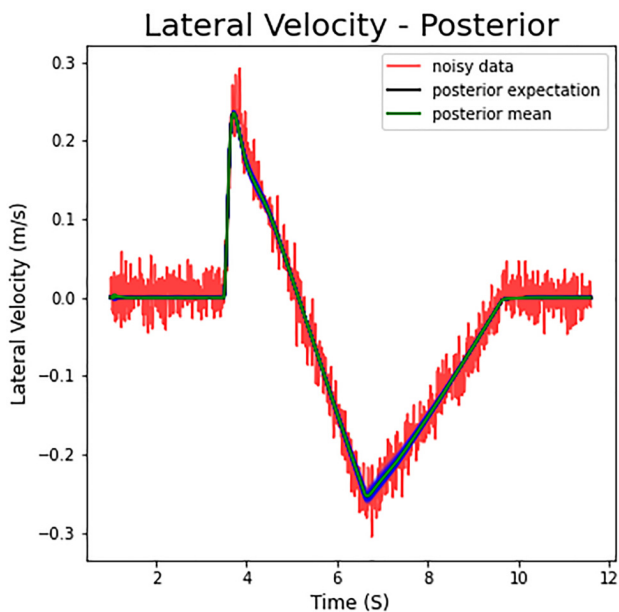

**Fig. 29** Comparison of the lateral velocity response of the 8DOF model with the noisy data with parameters drawn from the posterior

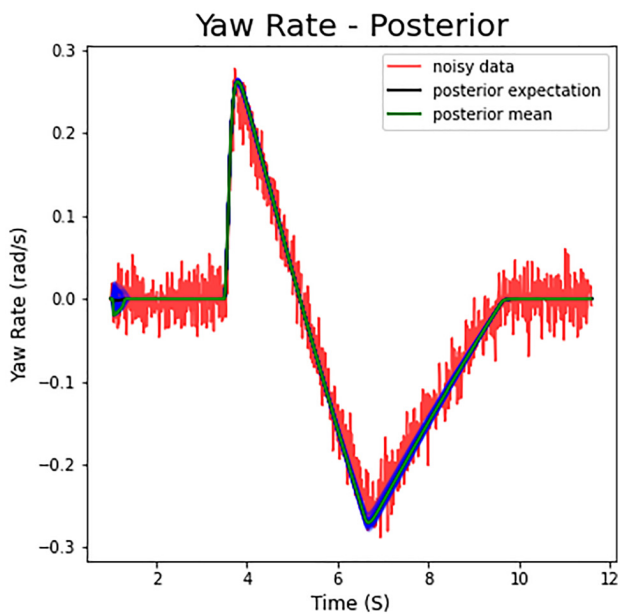

**Fig. 31** Comparison of the yaw rate response of the 8DOF model with the noisy data with parameters drawn from the posterior

## 4 Discussion

**4.1 Using Different Samplers.** There are multiple samplers available to solve a Bayesian calibration problem. Herein, three of them are employed to understand (a) how close they are in terms of the posterior distributions produced and (b) how efficient they are. Besides SMC, we considered the Metropolis-Hastings sampler [34], which is one of the first algorithms employed for MCMC, and the No-U-Turn Sampler (NUTS) [35], which works using the Hamiltonian Monte Carlo algorithm [10].

The Metropolis-Hastings algorithm is an adaptation of a random walk with an acceptance/rejection rule to converge to the specified target distribution [30]. While simple to implement, it is known to not perform very well in high dimensional parameter spaces due to its random-walk style of exploration [36].

Hamiltonian Monte Carlo (HMC) is a MCMC algorithm that avoids the random walk behavior and sensitivity to correlated parameters that plague many MCMC methods. It accomplishes this taking a series of steps informed by first-order gradient information

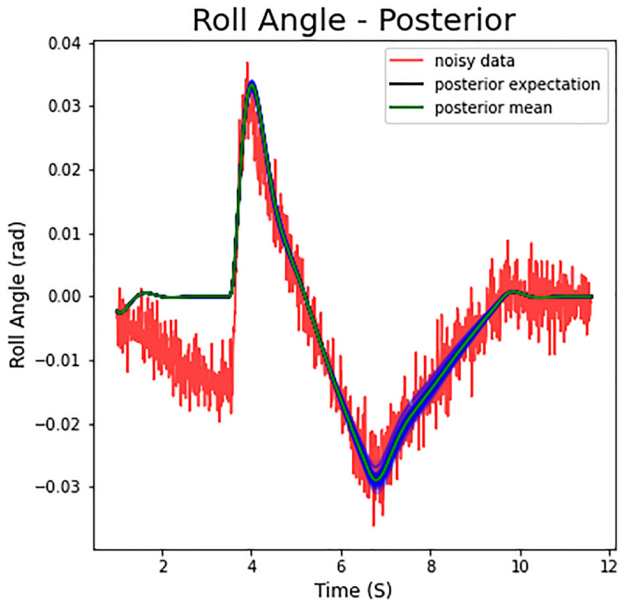

Fig. 32 Comparison of the roll angle response of the 8DOF model with the noisy data with parameters drawn from the posterior

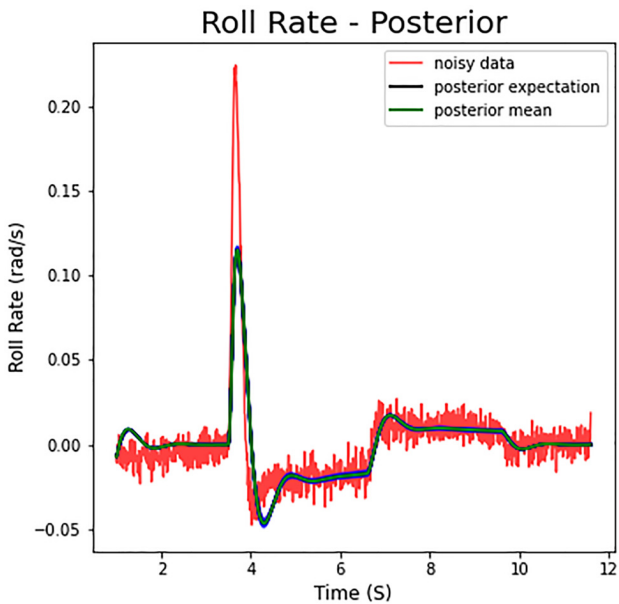

Fig. 33 Comparison of the roll rate response of the 8DOF model with the noisy data with parameters drawn from the posterior

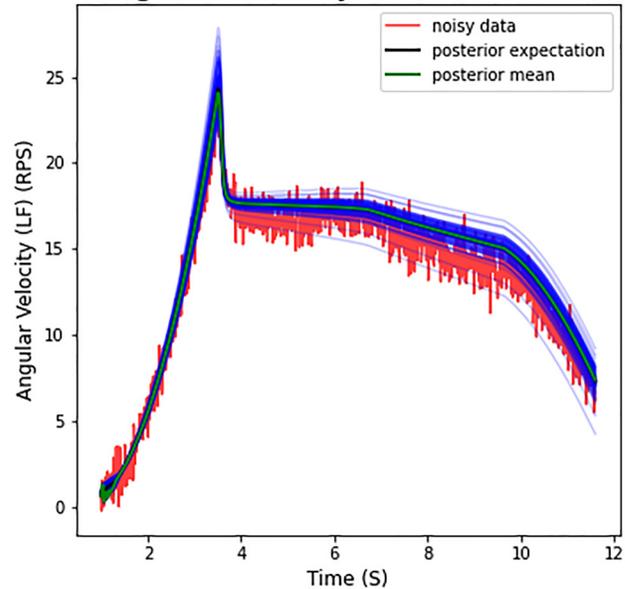

Fig. 34 Comparison of the angular velocity of wheel (left-front) response of the 8DOF model with the noisy data with parameters drawn from the posterior

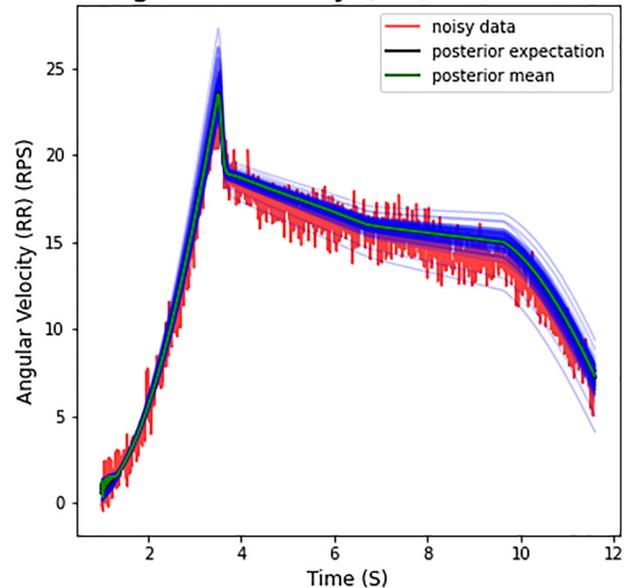

Fig. 35 Comparison of the angular velocity of wheel (right-rear) response of the 8DOF model with the noisy data with parameters drawn from the posterior

Table 10 Run time comparison

| Model | Simulation time (s) | Run time (s) |
| --- | --- | --- |
| 8-DOF | 11.5 | 0.37 ± 0.01 |
| Chrono::Vehicle | 11.5 | 4.82 ± 0.05 |

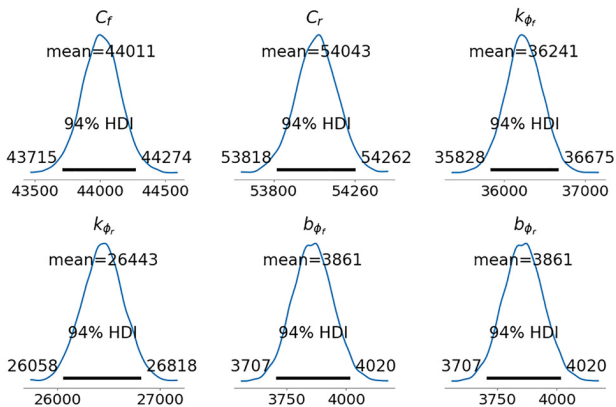

Fig. 36 Posterior produced using the NUTS sampler

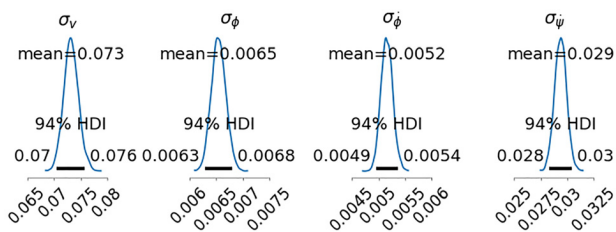

Fig. 37 Posterior of the noise produced using the NUTS sampler

[35]. This allows the HMC methods to explore high dimensional spaces much more efficiently by producing less correlated samples. NUTS is an extension of HMC. Among other things, it eliminates the need to set the step size and the number of steps, two crucial parameters for the computational efficiency of the HMC algorithm.

To gain insights into their performance, we perform the lateral dynamics calibration once again using all three samplers. The same priors are used for the parameters, and the sample acceptance probability is set to 0.9 for all samplers. NUTS and M-H differ from SMC in their need to have tuning samples. Tuning is the process of discarding the initial few samples drawn by the sampler since they typically do not belong to the converged target distribution [29]. The default number of tuning samples in PyMC is 500 and we use the same in this study. NUTS requires us to take the first order gradient of the likelihood functions; we do this via finite differences.

An effective sample size (ESS) per second metric is used to measure sampler efficiency. The split-$\hat{R}$ obtained for the different parameters is used to compare chain convergence and chain autocorrelation. Finally, it is insightful to check whether all three samplers arrive at a similar posterior albeit through different means.

The posterior obtained using the NUTS sampler can be seen in Figs. 36 and 37. The posterior is very similar to that produced by SMC in Figs. 13 and 14. However, the M-H algorithm does not produce a converged chain after 1000 draws (Fig. 38). We thus increase the number of draws per chain from 1000 to 10,000. Comparing results from 10,000 draws of M-H and 1000 draws of NUTS and SMC is still fair as the metric used for the comparison is scale free. With 10000 draws per chain, the M-H algorithm produces a converged posterior, see Figs. 39 and 40. This posterior is, again, almost identical to the posteriors produced using the NUTS and SMC sampler.

Table 11 Comparison of the Metropolis-Hastings (M-H), NUTS, and SMC samplers. T2C stands for time to completion for calibration when using a certain sampler (expressed in minutes).

| Sampler | T2C | Bulk-ESS | Bulk-ESS/s | $\hat{R}$ |
|---|---|---|---|---|
| NUTS | 567 | 6924 | 0.203 | 1.0011 |
| SMC | 123 | 7654 | 1.037 | 1.0003 |
| M-H | 439 | 5155 | 0.196 | 1.0039 |

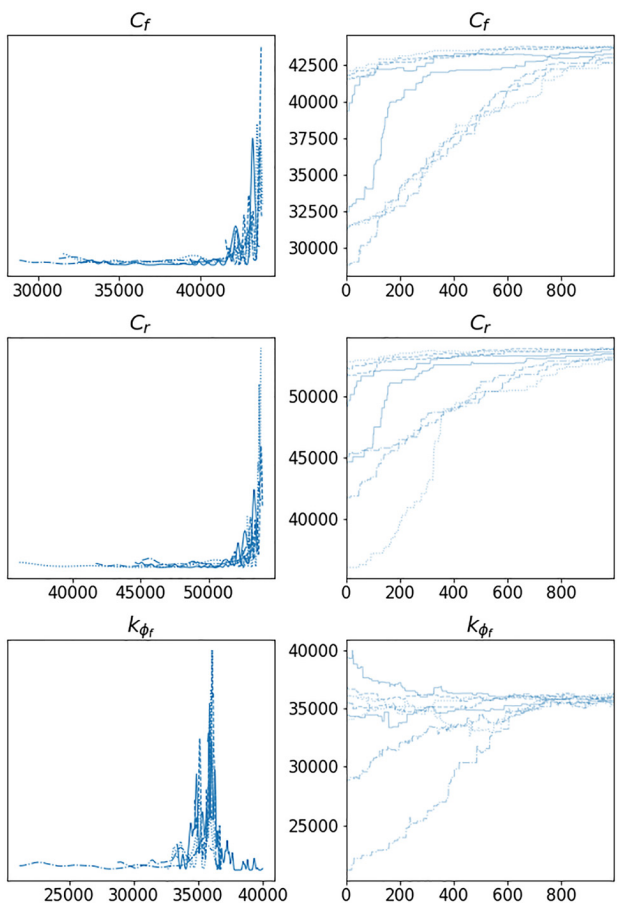

Fig. 38 Trace of 1000 draws produced using the Metropolis-Hastings sampler. The chain has not converged to the target distribution.

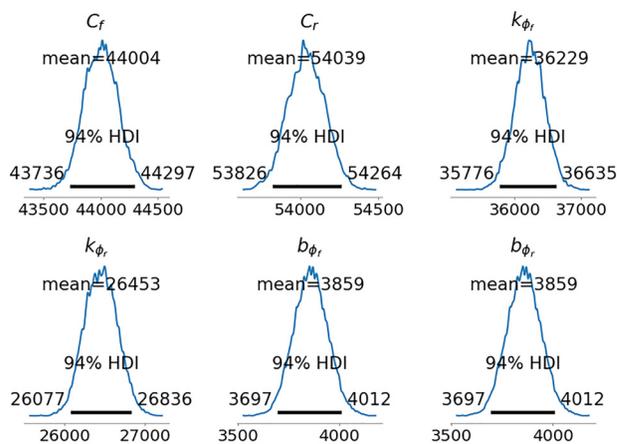

Fig. 39 Posterior of 10,000 draws produced using the Metropolis-Hastings sampler. The chain has now converged to the target distribution.

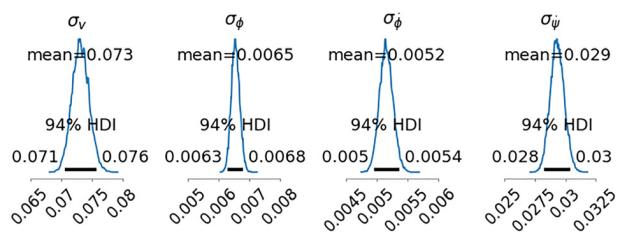

Fig. 40 Posterior of noise of converged chain using M-H sampler

Next, rather than comparing the ESS per second and split-$\hat{R}$ of all the parameters individually between the three methods, we take the average across all the parameters. Table 11 shows that SMC outperforms NUTS and M-H in all the metrics. Although we ran M-H for 10000 draws per chain, we only obtained a total of 5155 effective samples which shows that the M-H algorithm struggles to produce noncorrelated, independent samples. The NUTS algorithm does well to produce independent samples. However, as it requires the gradient of the likelihood for each draw, it is severely slowed down. The performance of NUTS could thus significantly improve if the model was implemented in a way that allows for techniques such as automatic differentiation (AD) to take derivatives rather than having to rely on finite differences.

### 4.2 Combining Rear and Front Roll Damping Coefficients.

This subsection highlights one typical scenario in which the Bayesian approach described cannot help. While calibrating the 8DOF model with the HMMWV model in the lateral direction, we had to reduce two of the model parameters, namely, the front ($b_{\phi_f}$) and rear ($b_{\phi_r}$) roll damping coefficient into a single vehicle roll damping coefficient $b_\phi$ by setting $b_{\phi f} = b_{\phi r} = 0.5 \, b_\phi$. This was done because initial calibration attempts showed that $b_{\phi_f}$ and $b_{\phi_r}$ are not uniquely identifiable using the HMMWV synthetic data available. This can be seen in the posterior pairwise plot between the two parameters (plotted after performing Kernel Density Estimation), see Fig. 41, as well as in the marginal posterior distribution which is not very sharp, see Fig. 42. Kernel Density Estimation (KDE) is a nonparametric way to estimate the probability density of a random variable based on kernels as weights. In Fig. 41, we provide a cross section of the joint—probability distribution of our two parameters where the more yellowish the hue, the higher the probability.

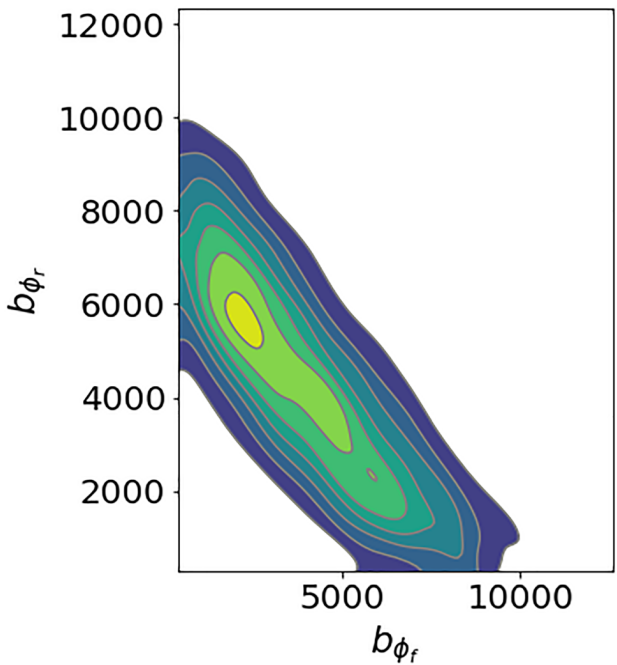

Fig. 41 Pairwise KDE plot between $b_{\phi_f}$ and $b_{\phi_r}$ shows nonidentifiability

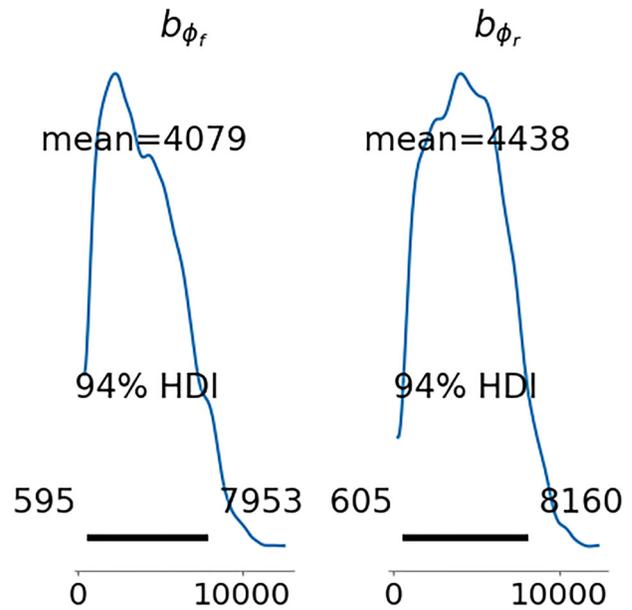

Fig. 42 Posterior of $b_{\phi_f}$ and $b_{\phi_r}$ not very sharp due to nonidentifiability

This can be explained using Eq. (7) where $b_{\phi_f}$ and $b_{\phi_r}$ occur together, as a summation, to provide the coefficient to the roll rate. Although in the vertical load transfer equations (Eqs. (10)), $b_{\phi_f}$ and $b_{\phi_r}$ occur independently, their low magnitude when combined with the roll rate is almost negligible in comparison with the rest of the terms. This also explains why the problem is not seen with $k_{\phi_f}$ and $k_{\phi_r}$, which have a sufficiently distinguishable effect on the vertical forces. This served as a reminder of a classical scenario in which the Bayesian calibration is not capable of providing accurate estimation of each individual parameter—when several parameters are nonidentifiable based on the data. Such kind of nonidentifiability would be manifested in the strong correlation between certain dimensions of the joint posterior distribution as illustrated in Fig. 41.

### 5 Conclusion and Future Work

This contribution outlines a Bayesian methodology to calibrate models used in robotics and autonomous vehicle applications, wherein there is a need for expeditious models to estimate state, plan motion, or determine control policies. The calibration approach discussed: (i) is robust in that it handles noisy data; and, (ii) it provides levels of confidence in the model parameters obtained via probability distributions. In other words, the approach does not produce one parameter set, but rather an infinite number of possible parameter sets, each labeled by a probability of explaining the experimental data observed. We relied on a Sequential Monte Carlo technique to draw samples from the joint posterior distribution of the model parameters, a process that in combination with classical Monte Carlo estimation opens the door to predicting with quantified uncertainties other quantities of interest tied to the dynamic evolution of the simplified model, e.g., most likely trajectory, most likely speed evolution.

Looking ahead, we plan to apply the same Bayesian inference framework to terramechanics calibration, wherein the goal is to use a high-fidelity approach to generate data subsequently used to calibrate a lower-fidelity but faster terramechanics model. Two cases are of interest—using a continuum representation of the terrain [37] to calibrate a lower-fidelity Chrono terrain [38,39]; or using a fully resolved, discrete representation of the terrain [40], to produce calibration data for a lower-fidelity continuum representation of the terrain [37]. Second, we plan to employ automatic differentiation to better gauge the potential of the Hamiltonian Monte Carlo approach, which is currently hindered by the lack of gradient information.

Finally, we are in the process of using a real vehicle to generate calibration data, instead of relying on synthetic data generated in Chrono.

The software used to generate the results reported herein is available on GitHub as open source and distributed under a permissive BSD3 license [19].

## References

[1] Choi, H., Crump, C., Duriez, C., Elmquist, A., Hager, G., Han, D., Hearl, F., et al., 2021, "On the Use of Simulation in Robotics: Opportunities, Challenges, and Suggestions for Moving Forward," Proc. Natl. Acad. Sci. USA, **118**(1), p. e1907856118.
[2] Elmquist, A., Young, A., Mahajan, I., Fahey, K., Dashora, A., Ashokkumar, S., Caldararu, S., et al., 2022, "A Software Toolkit and Hardware Platform for Investigating and Comparing Robot Autonomy Algorithms in Simulation and Reality," Preprint arXiv:2206.06537.
[3] Abarbanel, H. D. I., Creveling, D. R., Farsian, R., and Kostuk, M., 2009, "Dynamical State and Parameter Estimation," SIAM J. Appl. Dyn. Syst., **8**(4), pp. 1341–1381.
[4] Uchida, T., Vyasarayani, C., Smart, M., and McPhee, J., 2014, "Parameter Identification for Multibody Systems Expressed in Differential-Algebraic Form," Multibody Syst. Dyn., **31**, pp. 393–403.
[5] Zhu, Y., Dopico, D., Sandu, C., and Sandu, A., 05 2015, "Dynamic Response Optimization of Complex Multibody Systems in a Penalty Formulation Using Adjoint Sensitivity," ASME J. Comput. Nonlinear Dyn., **10**(3), p. 031009.
[6] Nada, A. A., 2021, "Simplified Procedure of Sensitivity-Based Parameter Estimation of Multibody Systems With Experimental Validation," IFAC-PapersOnLine, **54**(14), pp. 84–89.
[7] Shehata, M., Elshami, M., Bai, Q., and Zhao, X., 2021, "Parameter Estimation for Multibody System Dynamic Model of Delta Robot From Experimental Data," IFAC-PapersOnLine, **54**(14), pp. 72–77.
[8] Kennedy, M. C., and O'Hagan, A., 2001, "Bayesian Calibration of Computer Models," J. R. Stat. Soc.: Ser. B (Stat. Methodol.), **63**(3), pp. 425–464.
[9] Andrieu, C., De Freitas, N., Doucet, A., and Jordan, M. I., 2003, "An Introduction to MCMC for Machine Learning," Mach. Learning, **50**(1), pp. 5–43.
[10] Neal, R., 2011, "MCMC Using Hamiltonian Dynamics," *Handbook of Markov Chain Monte Carlo*, S. Brooks, A. Gelman, G. Jones, and X.-L. Meng, eds., Chapman and Hall/CRC, Boca Raton, FL.
[11] Foreman-Mackey, D., Hogg, D. W., Lang, D., and Goodman, J., 2013, "Emcee: The MCMC Hammer," Publ. Astron. Soc. Pacific, **125**(925), p. 306.
[12] Doucet, A., Freitas, N. D., and Gordon, N., 2001, "An Introduction to Sequential Monte Carlo Methods," *Sequential Monte Carlo Methods in Practice*, Springer, Cham, Swizerland, pp. 3–14.
[13] Del Moral, P., Doucet, A., and Jasra, A., 2006, "Sequential Monte Carlo Samplers," J. R. Stat. Soc.: Ser. B (Stat. Methodology), **68**(3), pp. 411–436.
[14] Neal, R. M., 2001, "Annealed Importance Sampling," Stat. Comput., **11**(2), pp. 125–139.
[15] Salvatier, J., Wiecki, T. V., and Fonnesbeck, C., 2016, "Probabilistic Programming in Python Using PyMC3," PeerJ Comput. Sci., **2**, p. e55.
[16] Tasora, A., Serban, R., Mazhar, H., Pazouki, A., Melanz, D., Fleischmann, J., Taylor, M., Sugiyama, H., and Negrut, D., 2016, "Chrono: An Open Source Multi-Physics Dynamics Engine," *High Performance Computing in Science and Engineering—Lecture Notes in Computer Science*, T. Kozubek, ed., Springer International Publishing, pp. 19–49.
[17] Serban, R., Taylor, M., Negrut, D., and Tasora, A., 2019, "Chrono::Vehicle Template-Based Ground Vehicle Modeling and Simulation," Intl. J. Veh. Perform., **5**(1), pp. 18–39.
[18] Kumar, R., Carroll, C., Hartikainen, A., and Martin, O., 2019, "ArviZ: A Unified Library for Exploratory Analysis of Bayesian Models in Python," J. Open Source Software, **4**(33), p. 1143.
[19] Unjhawala, H., Zhang, R., Hu, W., Wu, J., Serban, R., Negrut, D., 2022, "Bayesian Calibration Models and Scripts," Simulation-Based Engineering Lab, University of Wisconsin-Madison, accessed Mar. 8, 2023, https://github.com/uwsbel/public-metadata/tree/master/2022/calibrationBayesianExamples
[20] Shim, T., and Ghike, C., 2007, "Understanding the Limitations of Different Vehicle Models for Roll Dynamics Studies," Veh. System Dynamics, **45**(3), pp. 191–216.
[21] Chen, S., Xiong, G., Chen, H., and Negrut, D., 2020, "MPC-Based Path Tracking With PID Speed Control for High-Speed Autonomous Vehicles Considering Time-Optimal Travel," J. Central South Univ., **27**(12), pp. 3702–3720.
[22] Chen, S., and Negrut, D., 2019, "A MATLAB® Implementation of a Set of Three Vehicle Dynamics Models," Simulation-Based Engineering Laboratory, University of Wisconsin-Madison, Madison, WI, Report No. TR-2019-04.
[23] Taylor, M., 2015, "Implementation and Validation of the Fiala Tire Model in Chrono," University of Wisconsin–Madison, Madison, WI, Report No. TR-2015-13.
[24] He, J., Crolla, D. A., Levesley, M. C., and Manning, W. J., 2004, "Integrated Active Steering and Variable Torque Distribution Control for Improving Vehicle Handling and Stability," SAE Trans., **113**, pp. 638–647.
[25] Miles, P., Hays, M., Smith, R., and Oates, W., 2015, "Bayesian Uncertainty Analysis of Finite Deformation Viscoelasticity," Mech. Mater., **91**, pp. 35–49.
[26] Miles, P. R., and Smith, R. C., 2019, "Parameter Estimation Using the Python Package Pymcmcstat," Proceedings of the 18th Python in Science Conference, Austin, TX, July 8–14 pp. 93–100.
[27] Gelman, A., and Rubin, D. B., 1992, "Inference From Iterative Simulation Using Multiple Sequences," Stat. Sci., **7**(4), pp. 457–472.
[28] Vehtari, A., Gelman, A., Simpson, D., Carpenter, B., and Bürkner, P.-C., 2021, "Rank-Normalization, Folding, and Localization: An Improved $\hat{R}$ for Assessing Convergence of MCMC (With Discussion)," Bayesian Anal., **16**(2), pp. 667–718.
[29] Robert, C. P., Casella, G., and Casella, G., 1999, *Monte Carlo Statistical Methods*, **2**, Springer, New York.
[30] Gelman, A., Carlin, J. B., Stern, H. S., and Rubin, D. B., 1995, *Bayesian Data Analysis*, Chapman and Hall/CRC, Boca Raton, FL.
[31] Flegal, J. M., Haran, M., and Jones, G. L., 2008, "Markov Chain Monte Carlo: Can we Trust the Third Significant Figure?," Stat. Sci., **23**(2), pp. 250–260.
[32] Anitescu, M., and Tasora, A., 2010, "An Iterative Approach for Cone Complementarity Problems for Nonsmooth Dynamics," Comput. Optim. Appl., **47**(2), pp. 207–235.
[33] Hairer, E., Lubich, C., and Wanner, G., 2006, *Geometric Numerical Integration: Structure-Preserving Algorithms for Ordinary Differential Equations*, Vol. 31, Springer Science & Business Media, Berlin, Germany.
[34] Hastings, W. K., 1970, "Monte Carlo Sampling Methods Using Markov Chains and Their Applications," Biometrika, **57**(1), pp. 97–109.
[35] Hoffman, M. D., and Gelman, A., 2014, "The No-U-Turn Sampler: Adaptively Setting Path Lengths in Hamiltonian Monte Carlo," J. Mach. Learn. Res., **15**(47), pp. 1593–1623.
[36] Betancourt, M., 2017, "A Conceptual Introduction to Hamiltonian Monte Carlo," preprint arXiv:1701.02434.
[37] Hu, W., Zhou, Z., Chandler, S., Apostolopoulos, D., Kamrin, K., Serban, R., and Negrut, D., 2022, "Traction Control Design for Off-Road Mobility Using an SPH-DAE co-Simulation Framework," Multibody Syst. Dyn., **55**, pp. 165–188.
[38] Tasora, A., Mangoni, D., Negrut, D., Serban, R., and Jayakumar, P., 2019, "Deformable Soil With Adaptive Level of Detail for Tracked and Wheeled Vehicles," Int. J. Veh. Performance, **5**(1), pp. 60–76.
[39] Serban, R., Taves, J., and Zhou, Z., 2022, "Real-Time Simulation of Ground Vehicles on Deformable Terrain," ASME Paper No. DETC2022-89470.
[40] Fang, L., Zhang, R., Vanden Heuvel, C., Serban, R., and Negrut, D., 2021, "Chrono::GPU: An Open-Source Simulation Package for Granular Dynamics Using the Discrete Element Method," Processes, **9**(10), p. 1813.